# An Instrument to Evaluate Governance Proposals: AI Policy Analysis at Scale

**Paulo Carvão[1], Isabel Adler[2], Claudio Mayrink Verdun[3], Jeffrey Zhou[4]**

---

**"In our judgments we should be careful not to attribute more weight to things than they have."**
Jakob Bernoulli, *Ars Conjectandi* (published posthumously, 1713)

# Abstract

This paper introduces a policy analysis framework designed to support systematic, transparent assessment of artificial intelligence (AI) governance proposals in a rapidly evolving and contested regulatory landscape. AI policy debates often collapse into binary positions that obscure underlying tradeoffs and normative assumptions. The framework structures policy analysis around multiple policy attributes, allowing users to surface priorities and tensions without prescribing outcomes. The research adopts a mixed-methods approach that integrates qualitative insights from subject matter experts with computational text analysis to inform the design of policy attribute indices and rubrics. The resulting approach quantifies the relative emphasis of different policy objectives and presents them through comparative visualizations that support interpretability and cross-policy comparison. The paper also examines the use of commercial large language models for rubric-based policy analysis, benchmarking their outputs against a domain-trained rubric-calibrated model with explicitly defined analytical assumptions. Rather than assessing policy effectiveness or desirability, the framework focuses on relevance and alignment across attributes. By making analytical assumptions explicit, including attribute selection, rubric construction, and weighting schemes, the framework enables users to evaluate whether its embedded priorities align with the users' own normative commitments. The approach is jurisdiction-agnostic and intended to support policymakers, analysts, and researchers navigating complex AI governance environments.

The framework makes three contributions: (1) it operationalizes multidimensional policy assessment through empirically grounded rubrics that surface tradeoffs rather than resolving them; (2) it develops a transparent hybrid methodology combining feedback from subject-matter experts with computational validation; and (3) it demonstrates how domain-trained rubric-calibrated models can be used as a benchmark for comparing different general-purpose large language models. These methodological advances enable more systematic, reproducible policy comparison while maintaining transparency about embedded normative choices.

**Biographical Notes**

**Paulo Carvão** is a Senior Fellow at Harvard Kennedy School's Mossavar-Rahmani Center for Business and Government, focusing on Tech Policy and AI regulation. A former IBM executive, he advises tech startups and serves as Entrepreneur-in-Residence at the Harvard Institute of Politics Governance Lab. Mr. Carvão has held fellowships at Harvard's Advanced Leadership Initiative and Safra Center for Ethics.

**Isabel Adler** is a dual Harvard Kennedy School master's in public policy (MPP) and Tuck School of Business at Dartmouth (MBA) candidate. Before graduate school, she worked at the Knight First Amendment Institute, which safeguards public discourse in the digital age.

**Jeffrey Zhou** is an A.B. candidate in Computer Science and Economics at Harvard College with industry experience at Amazon Web Services and venture-backed startups. He was a former research assistant at the Harvard Kempner Institute, working on intelligent vision algorithms for robotics and medical assistance tools for physicians. He has also developed AI tools for policy research and analysis.

**Claudio Mayrink Verdun** is a Postdoctoral Research Scholar at Harvard School of Engineering and Applied Sciences, working on the mathematical foundations of responsible AI. He researches frameworks, algorithms, and theoretical guarantees for safe and equitable AI, with a focus on inference-time alignment, fairness, and interpretability. He works with policymakers on governance, including contributing to G20 summit policy discussions, and serves as a Technical Development Lead for the Alignment Research Engineer Accelerator in Cambridge, MA. He holds a Ph.D. (summa cum laude) in mathematics and electrical engineering from the Technical University of Munich.

[1] Corresponding author – Mossavar-Rahmani Center for Business and Government (paulo_carvao@harvard.edu, pcarvao65@gmail.com)

[2] Harvard Kennedy School of Government (isabeladler@hks.harvard.edu)

[3] School of Engineering and Applied Sciences., Harvard University (claudioverdun@g.harvard.edu,claudioverdun@gmail.com)

[4] Faculty of Arts and Sciences, Harvard University (jeffreyzhou@college.harvard.edu)

## Table of Contents

## List of Figures

## List of Tables

# Introduction

Policymakers face an impossible task. Artificial intelligence, which is increasingly recognized as a general-purpose technology on par with steam power, electricity, and the internet (Brynjolfsson & McAfee, 2016; Calvino et al., 2025; Trajtenberg, 2019), is reshaping every sector of the economy. As generative AI diffuses rapidly (Bick et al., 2024), legislators confront a vast quantity of proposed bills, competing expert testimony, and a fragmented regulatory landscape with no standardized method to compare policies. The framework and the associated instrument (tool) presented here address that problem: a systematic, transparent method for evaluating AI governance proposals that cuts through ideological noise and surfaces the tradeoffs embedded in any given policy.

Faced with this complexity, many policymakers and analysts have begun using commercial large language models to interpret and even score AI legislation. These tools offer speed and convenience, but their outputs vary across prompts and models, and their embedded normative assumptions are rarely visible. Policy evaluation, however, demands reproducibility, structured criteria, and transparency about value choices. The framework presented here does that.

Scholars and practitioners are now debating AI's effects on productivity, labor market restructuring, and its capacity to address global challenges, including climate change, global health, food security, education, and accelerating scientific discovery (Birhane et al., 2023; Brynjolfsson et al., 2025; Eloundou et al., 2024; Filimonovic et al., 2025; Sharma, 2024). A parallel line of inquiry focuses on longer-term risks, including the potential for advanced AI systems to pose catastrophic or existential threats. The International AI Safety Report provides an in-depth assessment of risks posed by general-purpose AI systems (Bengio, Mindermann, et al., 2025; Bengio, 2026). The Singapore Consensus on Global AI Safety Research Priorities identifies key research directions for building trustworthy, reliable, and secure AI ecosystems (Bengio, Tegmark, et al., 2025). Moreover, academic reviews have systematically catalogued the risks associated with artificial general intelligence (McLean et al., 2023) (Hendrycks et al., 2023). Perhaps most striking is the Statement on AI Risk, signed by leading AI scientists and executives, which asserts that "Mitigating the risk of extinction from AI should be a global priority alongside other societal-scale risks such as pandemics and nuclear war." (Center for AI Safety, 2023)

The debate on how to govern such an important technology quickly divides between those who want to remove all barriers to development and those who prioritize safety. Politically, in the U.S., the situation is framed as a winner-takes-all race towards economic and geopolitical supremacy. Public, private, and sovereign investments soar, and AI becomes an engine for short-term economic growth. Betting against it can be seen as unpatriotic, foolish, or an impediment to national progress, marginalizing skeptical or cautionary voices. (Carvão, Jeloka, et al., 2025) On one side, advocates of a laissez-faire approach point to how the policies in the 1990s that shielded the Internet from regulation led to the creation of the modern information and electronic commerce age. (Chander, 2014) On the other hand, those advancing a more precautionary approach associate the current mental health crisis and democratic backsliding phenomena with the liability protections and lack of regulation of social media platforms over the last two decades. (Bradford, 2023; Haidt, 2024; Persily & Tucker, 2020)

Which side is "right"? Can a systematic framework make the value judgments embedded in policy analysis more transparent and open to contestation? Is value-neutral policy analysis even possible? The answer is no. Every framework embeds normative choices in what it measures, how it weighs components, and how it presents results (Fischer & Forester, 1993; Robert & Zeckhauser, 2011). The question is not whether these choices exist, but whether they are visible, documented, and contestable. This paper proposes a governance analysis model and an AI-powered analysis tool that will give policymakers, staffers, analysts, and researchers a systematic method for answering these questions themselves.

Existing policy analysis approaches face three issues: qualitative frameworks provide depth but do not scale; quantitative systems enable comparison but obscure trade-offs; computational methods scale but operate at a high level of abstraction. At the same time, AI governance can benefit from three simultaneous capabilities: systematic comparison across hundreds of policies, transparent application of substantive criteria, and visualization of multidimensional trade-offs across different policy aspects or goals. No existing framework combines these features while documenting its normative commitments explicitly. This paper's framework addresses that gap through three innovations, namely, rubric-based multidimensional scoring that makes tradeoffs visible, a hybrid human-AI validation methodology that grounds computational analysis in expert judgment, and transparent documentation of design choices that enables users to evaluate alignment with their own priorities.

Ultimately, policymaking is a political act. It is normative and opinionated. The framework introduced here neither judges, endorses, nor rejects policy proposals. Instead, it quantifies how strongly a given policy aligns with different governance priorities and presents those alignments through intuitive visualizations and scores. The scoring system is transparent to the user, and the tool features indices created through rigorous quantitative and qualitative research. With the tool, a legislator can quickly identify whether a bill emphasizes consumer protection or export controls, whether it prioritizes workforce development or cybersecurity. Transparency about the development of the indices' rubrics allows the user to decide whether those priorities, and their relative weighting, align with their own.

Policies can work in two different ways: they can restrict activities (which is what one usually thinks of as regulation), or they can enable activities by creating incentives or setting up the right conditions for certain actions, rights, or innovations to happen. Additionally, this work assumes that a policy can address one or more goals—which this paper calls attributes—such as safety, national security, environmental concerns, intellectual property, antitrust, and the responsible use of technology, among others. The attributes selected for this work are safety & security, national security, civil & human rights, antitrust, and industrial policy (economic resilience). A policymaker, in defining the objectives of the policy, chooses the mix and relative importance of these attributes in the context of their normative framework and what is desired and politically viable in their jurisdiction. (Carvão, Ancheva, et al., 2025)

This article includes a brief history of regulatory theory, an explanation of our research approach, and examples of the framework and tool's applications to AI policy examples. The work also examines the use of commercial LLMs for rubric-based policy analysis and compares their output against our methodology to assess interpretability in automatic policy scoring. The tool is available at the website HKSCarvaoGroup.github.io.

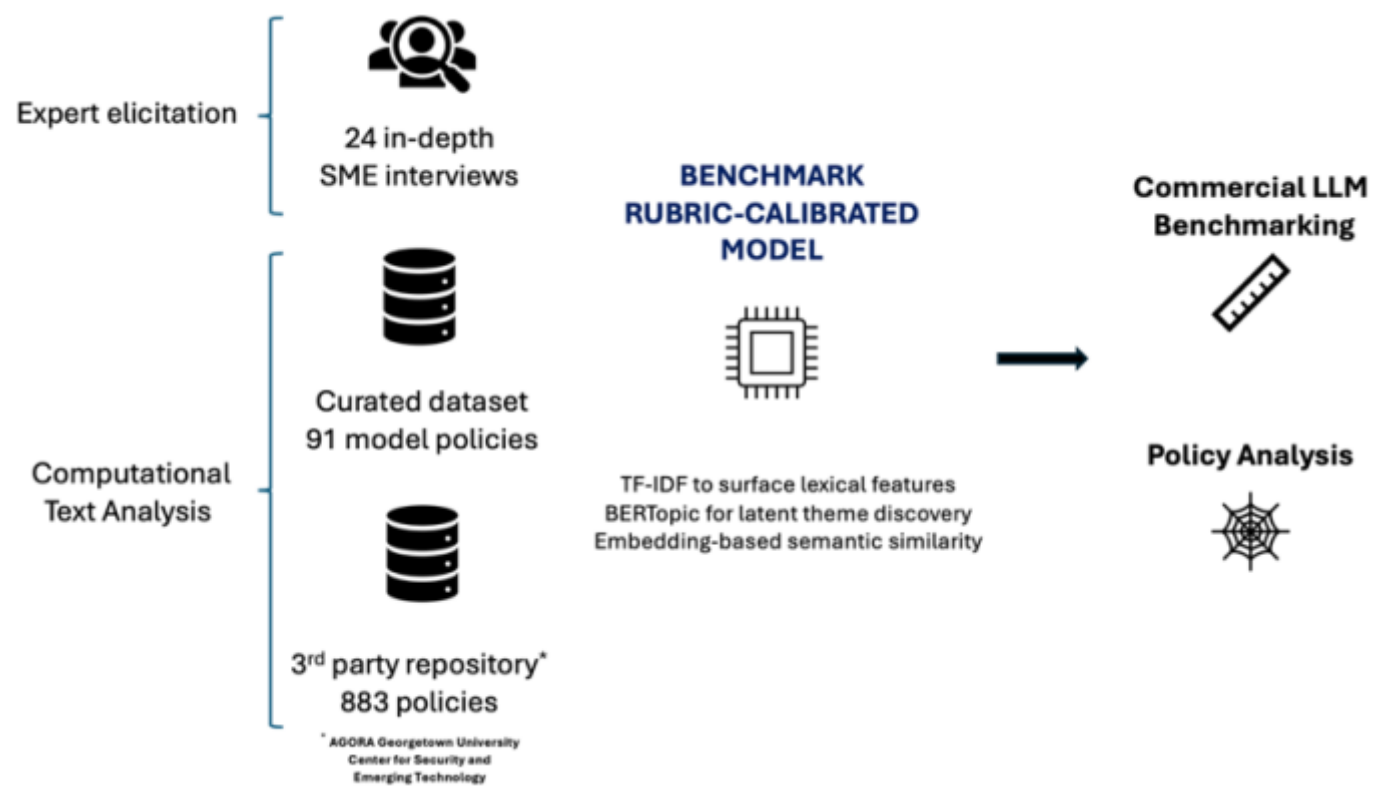

*Figure 1 Transparent AI Policy Analysis at Scale*

## Background on Regulatory Theory and Policy Analysis

Industrialization, market concentration, and recurring economic crises prompted modern regulation in the U.S. Early federal interventions focused on railroads, banking, and food safety, reflecting concerns that unregulated markets could produce systemic risks and asymmetric harms. Progressive Era reforms introduced independent regulatory commissions such as the Interstate Commerce Commission (1887) and the Federal Trade Commission (1914), establishing a model that combined technical expertise with political insulation. (Breyer, 1984; Glaeser & Shleifer, 2003) The New Deal marked an expansion of regulatory authority. Financial market oversight, labor protections, and social insurance programs institutionalized the idea that government intervention could stabilize markets and protect the public from economic shocks. Later, after the Second World War, policy makers extended regulations into communications, energy, and transportation, often focusing on natural monopoly theory or public utility frameworks.

During the 1970s, growing criticism against regulatory expansion emerged, resulting in the deregulation of sectors such as airlines, trucking, and telecommunications. (Derthick et al., 1985) At the same time, the legal and economic aspects gained prominence since policymakers started to focus on incentives, cost-benefit analysis, and market effectiveness. Regulatory policies began to shift, limiting government action to situations with evident and measurable market failures.

This evolution created a hybrid system in the U.S. in which market solutions are the default, with regulation serving as a correction rather than a planning tool. This balance continues to influence current discussions, including those about AI.

### Why and Why Not to Use Regulation

Regulation serves several core functions within a market economy. It addresses coordination failures, mitigates externalities, corrects information asymmetries, and manages systemic risks. In sectors characterized by rapid technological change, regulation can also create predictability that supports long-term investment.

Regulatory theory has also sought to systematize how regulation operates and why it succeeds or fails. Stephen Breyer, in *Regulation and Its Reform*, framed regulation as a distinct discipline, organizing regulatory programs by their purposes and recurring problems, and drawing on economics, law, and administration to extract lessons applicable across sectors and reform efforts (Breyer, 1984). For example, regulation can play an enabling role, and some scholars see well-designed regulation as a stimulus for innovative activity. Intellectual property regimes, standard-setting processes, and public infrastructure investments illustrate how regulatory structures can expand markets rather than restrict them. In technology sectors, public funding and procurement often shape innovation trajectories long before commercial viability emerges. (Ashford & Heaton, 1976)

In the context of AI, these rationales appear frequently. Concerns about safety, discrimination, national security, and infrastructure concentration reflect conditions where market incentives alone may not, in the first moment, when the technology is still being developed, internalize broader social costs. Regulation offers a mechanism to align private incentives with public objectives, provided design and enforcement remain proportionate (Carvão, 2024; Castañeira et al., 2025).

Despite its benefits, regulation carries risk. Poorly designed rules can suppress innovation, entrench incumbents, and impose compliance burdens that outweigh social and economic gains. Overregulation can lead firms to shift activity to less-regulated jurisdictions or design around rules in ways that preserve risk and increase opacity. Besides that, regulatory lag poses a challenge in fast-moving sectors, where static rules struggle to keep pace with evolving technologies. This is particularly remarkable in emerging technologies, where prescriptive mandates risk freezing early assumptions into law, limiting experimentation and adaptive learning (Carvão, Atir, et al., 2025). These concerns explain calls for restraint, particularly when regulatory objectives lack consensus or empirical grounding. Regulatory failure occurs when intervention fails to achieve its stated objectives or produces net social harm.

George Stigler's *The Theory of Economic Regulation* presents a more fundamental critique. Stigler argues that regulation often reflects the interests of regulated entities rather than the public. Firms with concentrated stakes possess stronger incentives to influence rulemaking than dispersed consumers, leading to outcomes that preserve market power under the guise of oversight (Stigler, 1971). Stigler's analysis frames regulatory capture as a predictable outcome rather than an anomaly. Because the interests of the voting public are diffuse, large industry players can use their clout to break through the noisiness of the electorate's needs and forefront industry priorities. Capture need not involve corruption; it often operates through expertise asymmetries, revolving doors, and agenda-setting power. Cass Sunstein, in *Paradoxes of the Regulatory State*, highlights a related tension. Regulation frequently aims to correct cognitive and behavioral biases, yet regulators themselves remain subject to institutional and psychological constraints. Risk perception, availability bias, and political salience can distort priority-setting, leading to overreaction in some domains and neglect in others (Sunstein, 1990).

## The Unique Challenge of Regulating Emerging Technology

Emerging technologies present distinct challenges that exacerbate the tensions identified in traditional regulatory theory. One of these challenges is the Collingridge Dilemma (Collingridge, 1980), which states that in a technology's early stages, intervention is possible, but impacts are uncertain; once impacts become clear, the technology is entrenched, and change is costly. Policymakers must act amid uncertainty about capabilities, trajectories, and societal effects. The "pacing problem," a term describing the structural inability of legal institutions to keep up with technological change, compounds this challenge (Marchant et al., 2011; Moses, 2007).

More recently, scholars have applied these frameworks to AI governance, observing that the rapid rollout of generative AI has outpaced policy development, with many risks becoming apparent only after systems are already widely deployed (G'sell, 2024). Meanwhile, as (Lessig, 1999) argued, technology architecture itself functions as regulation, shaping behavior through design choices that may escape legislative scrutiny.

These dynamics explain both the proliferation of AI policy proposals and the difficulty of evaluating them. Indeed, traditional deliberative processes prove ill-equipped for the speed and complexity of the technology they seek to govern.

## The Current State of AI Regulation

Globally, there is a high volume of AI legislation and little consensus about best practices (Hilliard et al., 2026). Over the last few years, U.S. federal and state legislatures have introduced hundreds of AI-related bills addressing safety, discrimination, national security, labor, and innovation. The EU AI Act's risk-based classification system reflects an effort to structure proportionate obligations by risk level, aiming to balance fundamental-rights protection with innovation and internal-market objectives (De Gregorio & Dunn, 2022). Despite international concern about the creation and use of AI, the regulatory landscape remains fragmented, and policymakers, analysts, compliance officers, and the public confront an information overload problem (Allen et al., 2025; Jobin et al., 2019).

The large volume of proposals makes any analysis challenging and comparative evaluation difficult. Without explicit and clear tools for their inspection, tradeoffs among policy bills remain implicit rather than explicit. Comparative assessment across jurisdictions or policy attributes often depends on subjective interpretation, which limits the reproducibility of the analysis and findings within such proposals. Overlapping federal, state, and international rules create ambiguity around obligations, increasing legal risk and compliance costs. As a result, smaller firms may face disproportionate burdens, potentially reinforcing market concentration. Moreover, public debates often collapse into binary frames that obscure substantive design choices. This leads to a legitimacy gap, as regulatory discussions appear opaque, technical, and disconnected from lived experience. These dynamics, in turn, erode trust and slow policy learning, reinforcing polarization rather than convergence.

## Using AI to Address Regulatory Complexity

Advances in natural language processing (NLP) enable systematic analysis of large policy corpora, supporting classification, comparison, and trend identification at scale. AI helps analyze regulations by highlighting policy priorities. While it does not resolve value disagreements, it can be used to clarify them and separate data from opinions, allowing for better-informed deliberation by making policy tradeoffs more visible and comparable. (Sanders & Schneier, 2025)

The methodology introduced here draws on two distinct intellectual traditions. The first is professional policy analysis, which treats evaluation as a client-oriented practice structured around problem definition, creation of alternative policy options, and criteria-based assessment (Bardach & Patashnik, 2023; Weimer & Vining, 2017). This tradition emphasizes political feasibility and implementation capacity, typically culminating in cost-benefit analysis or the systematic weighing of tradeoffs. However, it assumes well-defined problems, stable causal models, and identifiable alternatives, conditions that AI's rapid evolution does not meet.

The second tradition is computational text analysis. Justin Grimmer and Brandon M. Steward established foundational principles for automating the analysis of political texts, emphasizing that such methods augment rather than replace human judgment and require extensive, problem-specific validation (Grimmer & Stewart, 2013). After several developments in the field, Zhong and colleagues surveyed the state of legal artificial intelligence, documenting how NLP can support tasks ranging from document classification and information extraction to judgment prediction and legal question answering (Zhong et al., 2020). Their review demonstrates that legal NLP has moved beyond proof-of-concept experiments toward practical applications capable of handling the complexity and specialized vocabulary characteristic of regulatory documents. Jin and Mihalcea describe how NLP enables distinct applications across the policy cycle: analyzing data for evidence-based policymaking, improving policy communication, investigating policy effects, and interpreting political phenomena to the public (Jin & Mihalcea, 2022).

Government agencies themselves have begun adopting these computational approaches. A comprehensive study found that by 2020, nearly half of major federal agencies had experimented with AI and machine learning for tasks including regulatory analysis, enforcement targeting, and public engagement (Engstrom et al., 2020). The report documents use cases across agencies, including the Securities and Exchange Commission and Food and Drug Administration. However, computational approaches typically operate at high levels of abstraction, identifying latent topics or sentiment without connecting findings to substantive policy criteria defined by domain experts.

Neither tradition alone adequately addresses the challenge of systematically comparing AI governance proposals across multiple policy dimensions while maintaining transparency about embedded normative assumptions. The framework introduced in this paper bridges both: it adopts from policy analysis the commitment to structured, criteria-based evaluation grounded in expert judgment, while borrowing from computational text analysis the capacity to process large document collections systematically. The result is a hybrid methodology that scales beyond the limitations of manual analysis while remaining anchored in theoretically motivated categories that domain experts can scrutinize and contest.

## Project Approach and Scope

This work focuses primarily on the United States. While most of the policy artifacts are from the U.S. (at both the Federal and State levels), the research team included a small number of canonical European Union and United Kingdom policies in the analysis, given their pioneering role in setting AI policy. Researchers in other countries can reproduce the method described here to create their own tools. See the section

*A* recent work by Pham and others at the ACLU and Brown University (Tuan Pham et al., 2026) applies computational methods to 1,804 AI and automated decision systems legislation introduced in U.S. states and the U.S. Congress between 2023 and April 2025. They focus on legislative structure and policy language diffusion, how language is copied from bill to bill. Their analysis emphasizes inter-bill similarity, reuse of model bill language, and intra-bill definitional dependencies, using topic modeling, text comparison, and graph-based methods to support legislative drafting and review. While their approach prioritizes structural clarity and lineage across a large corpus of state and federal bills, the present study focuses on attribute-level semantic alignment and rubric-based evaluation, enabling systematic comparison of policy objectives across documents rather than document similarity or definitional structure alone.

The following paragraphs detail the sequential steps of our methodology, illustrated in Figure 3. It ranges from policy attribute selection and rubric development through data collection and preprocessing to the applications of each analytical technique.
Implementation Steps below. However, it is important to note that the tool's scoring system is based on U.S. data, expertise, and perspective, and the model would require adaptation to reflect specific characteristics of different loci of analysis.

The objective of the tool is to read the text of an AI policy artifact and quantify its relevance across different policy areas (attributes). The paper *Governance at a Crossroads* (Carvão, Ancheva, et al., 2025) identified ten attributes of interest for AI policies: antitrust, safety & security, responsible and ethical AI, national security, industrial policy, public interest (retitled civil & human rights in this paper), labor considerations, copyright implications, international collaboration, and elections. In the context of this work, an AI policy is anything that controls the creation, use, and diffusion of AI. For scope limitation, the analysis focuses on five attributes: antitrust, safety & security, national security, industrial policy, and civil & human rights. This selection is not meant to rank attributes by importance; instead, it aims to keep the project's initial scope manageable while ensuring a variety of attributes are included in the analysis.

The tool uses rubrics to evaluate a policy's alignment with each attribute (see Appendix I). The framework applies different weights to sub-components within each rubric. Four considerations guided the rubrics' design: (1) capture and incorporate feedback from subject-matter experts; (2) clearly state assumptions to build trust; (3) ensure that the evaluation process is data-driven; and (4) use open-source software to enable future feature development and customization by users. As an example, below, one can find an overview of the Industrial Policy (Economic Resilience) attribute and its associated evaluation rubric.

**Industrial Policy (Economic Resilience)** refers to strategic measures that strengthen U.S. competitiveness and economic resilience in the global AI economy. It seeks to position the U.S. as a global leader in AI while supporting job creation and innovation across industries.

**Industrial Policy (Economic Resilience) Index Rubric**

a. **Workforce Development & Job Quality** (Weight 25%)
b. **R&D, Investment & Incentives** (Weight 20%)
c. **Domestic Manufacturing & Supply Chains** (Weight 20%)
d. **Infrastructure & Compute Capacity** (Weight 15%)
e. **Energy & Resource Capacity** (Weight 10%)
f. **Global Competitiveness** (Weight 10%)

One of the framework's primary contributions is the blend of qualitative and quantitative research methodologies. The research team performed in-depth interviews with Subject Matter Experts (SMEs) across the policy attribute areas and applied machine learning and AI-driven quantitative methods to a dataset of model policies.

The resulting AI-powered tool processes a given policy and produces an analysis of the artifact. This includes a summary, a visualization (spider graph) along the five predetermined policy attributes, policy excerpts aligned to each attribute, and detailed scoring criteria for each attribute (rubric scoring details). Figure 2 depicts these features as implemented in the tool prototype, available at HKSCarvaoGroup.github.io. This helps visualize the policy content alignment and relevancy to different attributes. It is important to note that the tool does not evaluate the policy's effects as positive or negative; such determinations are left to the user's normative judgment.

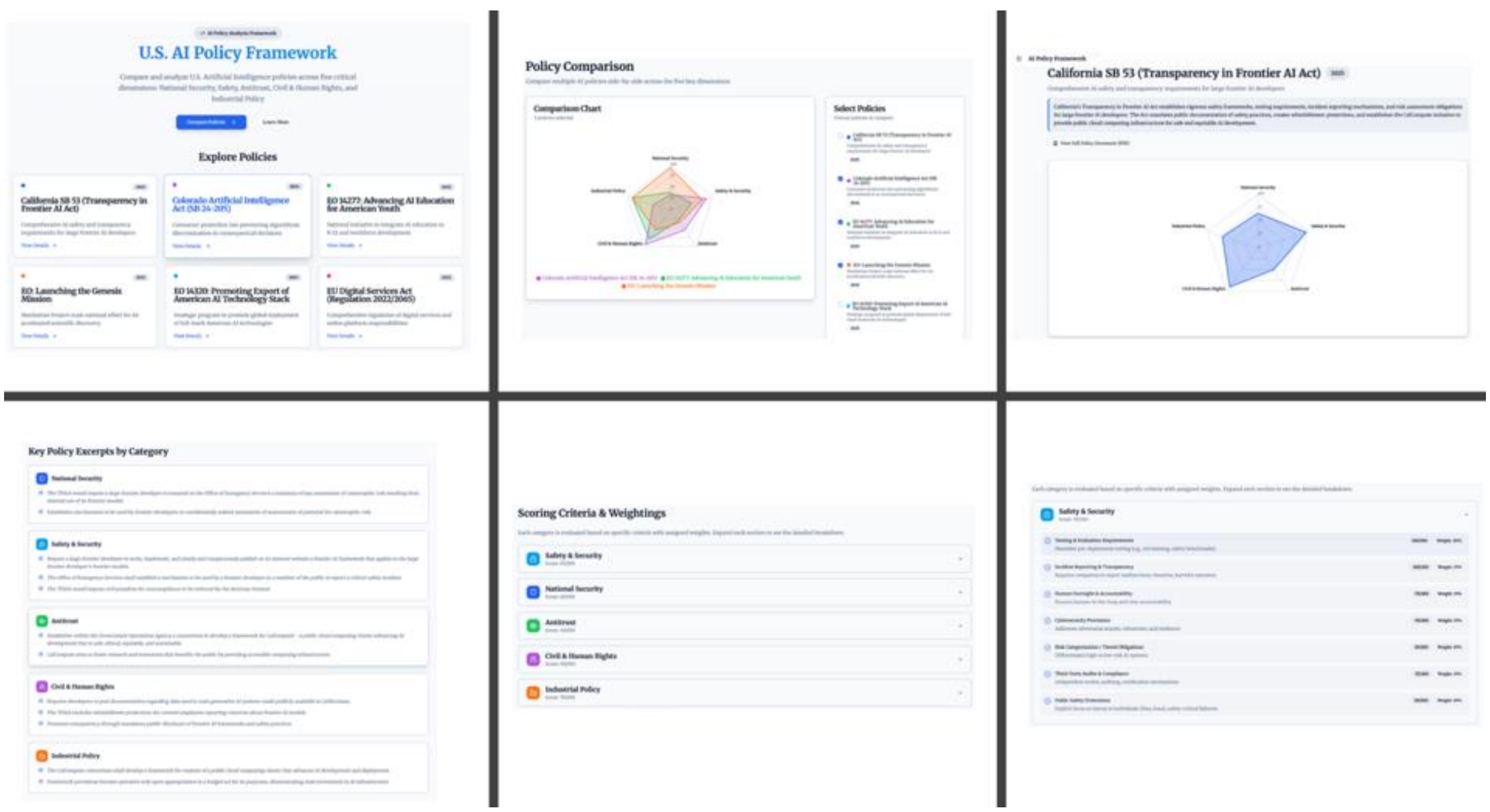

*Figure 2 Tool Prototype Snapshots*

There are various ways in which the tool can be used. A policy analyst might access the tool to identify similarities and differences between recent AI legislation introduced in two U.S. states. A compliance officer at an AI company might monitor emerging industrial policy regulation by inputting different versions of a bill into the tool, which will identify its key industrial policy components.

## The Framework's Normative Architecture

Policy analysis frameworks necessarily embed normative choices at multiple levels. Claims to the contrary obscure rather than eliminate value judgments (Fischer & Forester, 1993). This framework makes its normative architecture explicit across three levels.

First, attribute selection reflects prioritization. The choice to analyze Safety & Security, National Security, Antitrust, Civil & Human Rights, and Industrial Policy rather than Environmental Impact, Labor Rights, or International Cooperation carries a judgment about which dimensions of AI governance merit attention. This selection emerged from analysis of prevailing policy discourse and iterative consultation with domain experts, but it remains a choice. The selection is not designed to be comprehensive or optimal.

Second, rubric weights encode contested judgments about relative importance. Within each attribute, the framework assigns differential weights to sub-components. Testing and Evaluation receives a 25 percent weight in the Safety & Security rubric, while Whistleblower Protections receives a 10 percent weight. These weights reflect the observed emphasis in model policies and elicited expert judgment, but they are not value-neutral facts about the world. Different weighting schemes would produce different scores. As (Robert & Zeckhauser, 2011) demonstrate in the climate policy context, small differences in weights can produce large differences in conclusions.

Third, the framework privileges specific analytical moves. Treating overlap between attributes as informative rather than problematic, preferring multidimensional scoring over categorical classification, and measuring statutory language rather than implementation outcomes each reflects a methodological commitment with normative implications. The choice to score relevance and alignment rather than effectiveness or desirability is itself a normative choice about what policy analysis should accomplish.

The framework's contribution lies not in eliminating these commitments but in documenting them with sufficient specificity such that users can evaluate whether the framework's embedded priorities align with their own analytical needs. As Weimer argues (Weimer, 2005), "neutral competence" in policy analysis does not mean absence of values; it means transparency about values combined with systematic application of those values across cases. This framework aspires to neutral competence in that sense, i.e., consistent application of documented analytical choices, not analysis purified of normative content.

In fact, users are invited to disagree with attribute selection, contest rubric weights, and challenge methodological commitments. Such disagreement does not invalidate the framework; it demonstrates the framework's capacity to surface the value judgments that shape policy assessment. To facilitate this, the complete codebase, including the embedding pipeline, rubric definitions, weighting schemes, and preprocessing workflows, is publicly available at GitHub (see Appendix III). Researchers and practitioners can reproduce the methodology with their own qualitative assessments, substitute alternative rubrics and weights reflecting their normative priorities, or extend the framework to additional policy attributes and jurisdictions.

# Research Methods

## Data Collection and Classification

This work used two complementary datasets for model training and for data analysis.

First, the research team curated a "model policy dataset" and used it to train a reference rubric-calibrated model. This dataset contains 91 total policies. The selection of the model policies relied on domain expertise and insights from twenty-four qualitative interviews and included legislative acts from the U.S. executive branch, the U.S. Congress, and U.S. states. In addition, four European Union policies, one United Kingdom policy, and one international standard were used, given their relevance to the antitrust, industrial policy, and safety & security attributes. This model policy dataset, as well as structured data from the qualitative interviews, were the basis for the controlled rubric-calibrated model training.

Next, to test and validate the training results and ensure that the rubric-calibrated model properly classifies policies outside of the model policy corpus, the study sourced 883 AI governance documents from the Emerging Technology Observatory AGORA database at Georgetown University's Center for Security and Emerging Technology. The selection of this source rested on its frequent update cycle and data structure.

This set of 883 governance documents includes policies issued by the U.S. Congress, federal agencies, state governments, and local governments. It excludes AI governance documents from foreign governments and privately owned companies, given this study's U.S.- and public sector-centric scope. The team manually spot checked the tool's policy area classifications against AGORA's policy tagging to assess validity. AGORA contains 78 metadata tags indicating the content area of each policy, 29 of which map onto a policy attribute in this paper's framework.

## Qualitative Research

Twenty-four subject matter experts (SMEs) participated in structured interviews on the policy analysis framework and tool. Expert selection drew from the authors' personal networks and referrals from interviewees. Most experts are U.S.-based, and all are familiar with AI policy. Experts come from a range of professional backgrounds: six currently or recently worked in government, eleven currently work in academia or think tanks, four from the corporate sector, and three entrepreneurs working in AI or AI policy.

Each session lasted around 30 minutes, was recorded, and anonymized. All interviewees received preparatory materials that explained the framework and included an image and description of the tool, defined the selected policy attributes, and detailed the preliminary rubrics for each. These were the interview questions:

1. Do the axes make sense to you? Should we add any? Remove any? Change any?
2. Do the rubrics make sense to you? Should we remove any lines? Add any? Change any?
3. What is the most famous, representative, or influential policy in your area of expertise? This does not have to be AI-related.
4. Do the axis definitions make sense to you? Would you edit them at all? If so, how?
5. What keywords come to mind when thinking about the axis/axes of your area(s) of expertise?
6. Could you see yourself using this tool? Why or why not? If yes, how would you use it?
7. Could you see anyone else using this tool? Why or why not? If yes, how would they use it?

The expert interviews sought to confirm conceptual understandings of AI governance within each attribute area, gather feedback on tool design criteria, indices, and rubrics, and identify potential use cases and users.

Subject matter experts indicated that rubric weights felt too abstract to critique numerically, so the qualitative interviews did not prioritize generating quantitative inputs. Experts instead reviewed the rubrics to assess whether individual line items aligned with their understanding of each policy area and identified instances where the proposed prioritization diverged from that understanding. Subsequent weight adjustments incorporated signals from the quantitative analysis.

Several experts interviewed agreed that although AI tools reflect their creators' biases, the research team could mitigate these concerns by providing clear, comprehensive transparency disclaimers explaining policy assumptions and technical methods. An industrial policy expert pointed out that safety & security, national security, antitrust, and civil & human rights are policy goals, while industrial policy is a tool, which may not meet the definition of a policy attribute. Two other experts noted that "Industrial Policy" has become increasingly synonymous with protectionism in current political discourse. The research team changed the label to "Industrial Policy (Economic Resilience)" to better reflect economic resilience as the policy goal and to clarify the definition of industrial policy.

These exchanges highlighted the fundamental tension in policy analysis framework design discussed earlier, namely the impossibility of value-neutrality. Every decision about what to measure, how to weigh it, and how to present results embeds normative commitments. The framework addresses this not by claiming neutrality but by documenting design choices with sufficient transparency that users can evaluate alignment with their own priorities.

Policy analysts asked for a feature that maps policy provisions to rubric lines, making it easier to compare similar provisions across various AI governance policies. This feature was implemented in the prototype.

In addition to conducting interviews, the research team also carried out an online survey with 16 participants, some of whom were interview subjects, while others were either not interviewed or opted to stay anonymous, to validate the rubric and identify keywords. One industry expert, who responded to the survey but was not interviewed, noted that the tool would be more helpful in public policy roles than in corporate governance, as corporate governance is company- and industry-specific. An anonymous, self-identified national security expert encouraged the research team to remove the national security axis because of "unacceptable error rates in high-stakes domains" and the tension between transparent, trustworthy AI and the—as the respondent put it— "we can't tell you how" nature of national security policy. This tension underscores the need for users to understand the normative framework used in the analysis. Highlighting error rates emphasizes that AI should assist humans, especially in significant decisions, a normative choice itself.

### *Integrating Qualitative Insights into Framework Design*

The qualitative interviews played an important role in shaping the framework and the design choices embedded in the tool. A recurring theme across interviews was a concern about overlap among policy attributes. Several experts noted that Safety & Security, National Security, and Civil & Human Rights often intersect in practice, reflecting different lenses applied to similar underlying risks. While some SMEs encouraged minimizing overlap among axes, others considered this overlap to be a signal of the strength of the tool's analytical capabilities. Overlapping scores can reveal where a policy addresses multiple objectives simultaneously, while also highlighting tensions between protection-focused and market-oriented approaches. This feedback reinforced the decision to allow policies to score across multiple attributes and informed the tool's emphasis on comparative visualization rather than categorical assignment.

Public Interest, retitled Civil & Human Rights in this paper, generated a particularly rich discussion. Interviewees noted that public interest derives from human rights such as privacy, fairness, and due process. The new name clarified the focus and helped refine rubric variables. Experts advised against viewing this category as opposed to innovation, suggesting instead that democratic oversight supports sustainable technology use.

Interviewees also raised normative distinctions among attributes. Antitrust was viewed as less value-neutral than other categories, as it presupposes concern about concentration and market power. An industrial policy expert noted that reliance on multiple provisions might counteract the impact of any one provision; the multi-line rubrics of the tool, therefore, might incentivize over-intervention. Some experts suggested a conceptual proximity between antitrust and industrial policy, while others emphasized antitrust's unique legal lineage and enforcement logic. These discussions underscored the importance of clearly defining each attribute and avoiding implicit judgments about policy desirability within the scoring process. Users should rely on their experience and judgment to interpret the tool's output and use discretion when incorporating the output into policy analysis and policymaking.

Another strong signal from the interviews was the importance of implementation, enforcement, and institutional context. Experts warned that statutory language alone offers an incomplete picture of a policy's real-world effects. In both the private and public sectors, a policy's impact is dependent on funding, enforcement, and the implementation of each provision over multiple years—not only on the text of the policy. One SME went as far as suggesting adding an enforcement line to every rubric. As a reminder, the tool measures alignment rather than effectiveness.

Trust and usability emerged as common prerequisites for the adoption of the tool. Interviewees expressed greater confidence in the framework and tool when assumptions, weights, and scoring logic were made explicit and inspectable. Experts viewed features that surface relevant policy excerpts, map provisions directly to rubric variables, and expose areas of overlap as particularly valuable. Several experts also emphasized accessibility to non-expert users, noting the potential for the tool to function as an educational aid for policymakers, staffers, students, and the public.

Finally, experts questioned how to interpret quality and impact in the context of policy analysis. Legislative traction, political salience, and policy quality do not always align. As a result, interviewees encouraged distinguishing analytical structure from normative evaluation. This feedback directly shaped the paper's emphasis on transparency, interpretability, and the separation of descriptive analysis from prescriptive judgment.

Interview findings highlight that AI governance analysis is inherently multidimensional, with overlaps and trade-offs that are analytically informative rather than problematic, and that any computational tool must remain grounded in transparent assumptions and human judgment.

## Quantitative Research

This study employs a layered quantitative text analysis strategy designed to examine AI policy documents at multiple levels of semantic abstraction, building on works by Ha, Wang et al., and Pham et al. that also apply computational text analysis to AI policy legislation (Ha, 2025; Tuan Pham et al., 2026; Wang et al., 2025). This paper extends this methodology by implementing progressively sophisticated techniques and introducing a novel embedding-based rubric validation system. Our analysis employed three methods that operate at increasing levels of semantic sophistication. We note that these quantitative methods were executed in parallel to the qualitative analysis, enabling independent discussion and subsequent triangulation of findings.

The first layer of our methodology employs **Term Frequency-Inverse Document Frequency (**TF-IDF) weighting to identify distinctive terminology within individual policy documents relative to the broader corpus. Originally developed by (Sparck Jones, 1972), TF-IDF measures the importance of a term to a document by combining its local frequency with an inverse function of its occurrence across all documents, effectively filtering out ubiquitous terms while surfacing lexical features that distinguish particular documents or clusters.

The second layer moves beyond term-level analysis to identify latent thematic structures within the corpus through the use of **BERTopic** for latent theme discovery (Grootendorst, 2022). This neural topic modeling approach generates document embeddings using pretrained language models, unlike conventional topic models such as Latent Dirichlet Allocation (LDA), which rely on bag-of-words representations. It reduces dimensionality via UMAP (Uniform Manifold Approximation and Projection), clusters documents using HDBSCAN (Hierarchical Density-Based Spatial Clustering of Applications with Noise), and extracts interpretable topic representations through class-based TF-IDF (c-TF-IDF). UMAP preserves semantic relationships when projecting high-dimensional embeddings into clusterable space (Healy & McInnes, 2024), while HDBSCAN identifies document groupings based on density without requiring a pre-specified number of clusters (Campello et al., 2013). This architecture enables BERTopic to capture semantic relationships that transcend lexical overlap.

**Embedding-Based Semantic Similarity.** The third analytical layer provides direct semantic comparison across documents using dense vector representations. This work employs Tencent's `KaLM-Embedding-Gemma3-12B-2511` (Zhao et al., 2025), a state-of-the-art multilingual embedding model that achieved top performance on the Massive Text Embedding Benchmark (MTEB) as of January 2026. This approach identifies substantive alignment between documents that use different terminology and distinguishes documents that share vocabulary but diverge in semantic content.

### *Methodological Rationale and Relation to Prior Work*

The three methods form a progressive analytical hierarchy. TF-IDF identifies what terms are distinctive. BERTopic reveals how documents cluster into coherent themes. Semantic embeddings measure the similarity of a document's meaning to the constructed rubrics of each policy attribute. Each successive layer captures increasingly abstract semantic properties, and findings from one layer complement the others and strengthen confidence in identified patterns.

In a related effort, Ha's work applies computational text analysis to AI policy documents, examining 204 AI bills introduced in the U.S. Congress between 2021 and 2024 using LDA topic modeling and lexicon-based sentiment analysis. Our approach extends this work in several dimensions. Where Ha analyzed bill summaries of 10 to 50 words, we processed 91 complete policy documents using a chunking strategy that preserves full regulatory language. Additionally, we also tested the method against an external dataset of 883 policies. We replaced LDA's bag-of-words representation with BERTopic's contextual embeddings, enabling the discovery of semantically coherent topics even when vocabulary varies. Rather than coarse-grained sentiment polarity via the Bing lexicon (sentiment analysis categorizing words as either positive or negative), our embedding-based approach captures nuanced semantic content across 31 rubric variables, validated against commercial LLMs. Finally, our methodology integrates qualitative and quantitative analysis: insights from expert interviews shaped the rubric definitions and attribute weights, which computational text analysis then validated against the structure of actual policy documents.

A recent work by Pham and others at the ACLU and Brown University (Tuan Pham et al., 2026) applies computational methods to 1,804 AI and automated decision systems legislation introduced in U.S. states and the U.S. Congress between 2023 and April 2025. They focus on legislative structure and policy language diffusion, how language is copied from bill to bill. Their analysis emphasizes inter-bill similarity, reuse of model bill language, and intra-bill definitional dependencies, using topic modeling, text comparison, and graph-based methods to support legislative drafting and review. While their approach prioritizes structural clarity and lineage across a large corpus of state and federal bills, the present study focuses on attribute-level semantic alignment and rubric-based evaluation, enabling systematic comparison of policy objectives across documents rather than document similarity or definitional structure alone.

The following paragraphs detail the sequential steps of our methodology, illustrated in Figure 3. It ranges from policy attribute selection and rubric development through data collection and preprocessing to the applications of each analytical technique.

## Implementation Steps

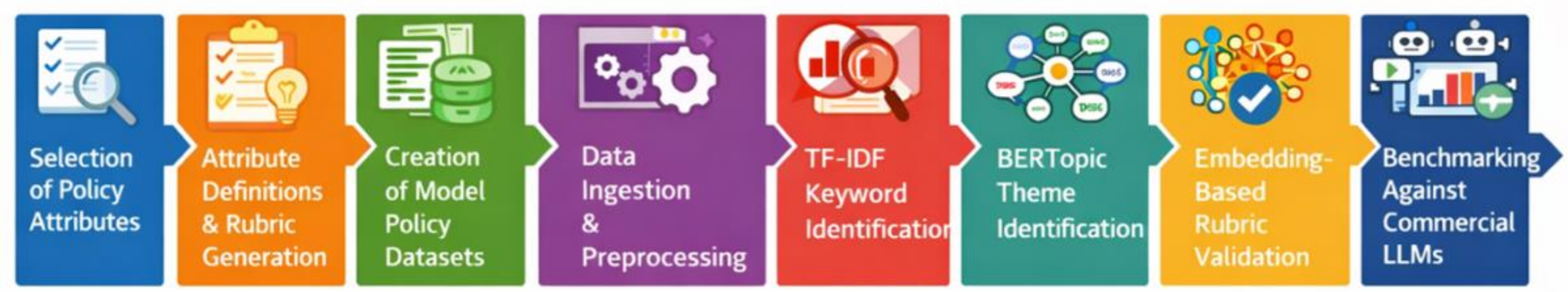

*Figure 3 Implementation Steps*

**Step 1: Selection of Policy Attributes**

The analysis draws on five policy attributes from the ten identified in *Governance at a Crossroads* (Carvão, Ancheva, et al., 2025). National Security, Safety & Security, Antitrust, Civil & Human Rights, and Industrial Policy feature prominently in current AI governance debates, appear across multiple jurisdictions, and emerged as salient during early qualitative interviews.

**Step 2: Attribute Definitions and Rubrics Generation**

Attribute definitions and rubrics evolved through iterative development informed by specialist interviews and topic extraction from model policies. Summary definitions for the five attributes underwent review and refinement during qualitative interviews with subject matter experts. Each rubric index included five to ten variables, balancing coverage against redundancy and scoring complexity. A literature review and cross-referencing with government datasets, including Congress.gov, NIST, and BIS, supported variable selection and weighting. Qualitative feedback informed rubric revisions and weight adjustments. The final system comprised 31 detailed rubric line-item definitions across the five policy areas, later used for granular semantic matching.

**Step 3: Creation of Model Policy Datasets**

The research team selected a set of model policies for each attribute area (n = 91 total), distributed as follows: Antitrust (n = 10), Industrial Policy (n = 28), National Security (n = 19), Civil & Human Rights (n = 14), and Safety & Security (n = 20). SMEs validated and refined the selection during qualitative interviews. The complete dataset listing appears in Appendix III.

**Step 4: Data Ingestion and Preprocessing**

A data preprocessing pipeline ingested and prepared policy artifacts for analysis, accommodating both structured sources via public APIs and unstructured documents.

The dataset includes structured documents, such as House and Senate bills, collected through the official Congress.gov API. Unstructured sources include executive orders and international legislation, such as the UK Digital Markets, Competition and Consumers Act (DMCCA), which U.S.-based APIs do not readily provide. The analysis sourced older statutes, including the Clayton Antitrust Act, through the GovInfo.gov API. The analysis accessed UK legislation through the Legislation.gov.uk API. For documents available only in PDF format, including the National Institute of Standards and Technology (NIST) AI Risk Management Framework, the workflow relied on the `pdfplumber` package for text extraction.

The preprocessing pipeline tokenized text using the `spaCy` library (Honnibal et al., 2020) after removing legal stop words. The process split each bill into 200-word chunks with 50-word overlap to preserve localized but substantively relevant passages for subsequent analysis rather than allowing unrelated sections of lengthy documents to dilute them. We opted for such a strategy since it would be hard to divide the bills into sections due to the variable size of sections in each bill.

Pooling 200-word chunks reduces dilution of localized provisions but introduces the risk that a high-similarity passage may amplify an isolated signal. Even so, consistent multidimensional patterns across methods and datasets indicate that the core finding does not hinge on pooling mechanics alone. A formal robustness analysis comparing alternative aggregation strategies would clarify the magnitude of this effect, but such testing falls outside the scope of this study.

**Step 5: TF-IDF Keyword Identification**

The analysis applied TF-IDF to each attribute-area cohort to identify the most frequent and distinctive keywords associated with each policy attribute. The workflow used the `scikit-learn` library, specifically the class `TfidfVectorizer`, for feature extraction and measured term distinctiveness through a one-vs-rest contrast scoring approach.

The TF-IDF analysis uses a cross-attribute approach designed to find terms that are genuinely distinctive to each policy area rather than merely frequent within individual documents. The analysis begins by loading all 91 model policies across all five policy areas and fitting a single TF-IDF vectorizer on the entire corpus. This ensures that the IDF component is calculated relative to all documents in the project. The vectorizer is configured to capture both single words and two-word phrases, which are often more meaningful in legal contexts. Terms must appear in at least 2 documents across the entire corpus to be included in the vocabulary, but cannot appear in more than 95 percent of documents to exclude ubiquitous terms.

For each policy area, the analysis computes a contrast score for every term by calculating the mean TF-IDF of the term across all documents within that policy area, then subtracting the mean TF-IDF across all documents outside that policy area. This approach naturally rewards terms that are both prevalent within the policy area and rare outside of it. A coverage filter requires that each term appear in at least 30 percent of documents within that policy area, preventing terms that appear in only one or two bills from dominating rankings while capping coverage at 95 percent to exclude terms too common to be meaningful differentiators.

**Step 6: BERTopic Theme Identification**

The analysis applied BERTopic modeling to each model policy cohort to identify thematic structures. The workflow ran BERTopic with a custom vectorizer that removed numeric tokens, such as years and section numbers, to isolate semantic policy language. KMeans clustering enforced exactly ten topics per policy area to support standardized comparison and align with the maximum number of rubric variables. Manual interpretation of each topic's keyword distribution produced descriptive labels, which informed validation of the rubric variables.

The analysis utilized guided BERTopic modeling to estimate topic relevance within individual documents. Topic keywords and descriptors functioned as seed words to bias the initial embedding space. Rather than relying on traditional LDA probabilities, the model generated document–topic distributions based on the proximity of document embeddings to seeded topic centroids. Ranking these distributions by average and cumulative strength provided a basis for validating rubric weights. At a higher level, the model treated policy attributes as predefined topics, producing a mapping of attribute relevance across the policy corpus.

**Step 7: Embedding-Based Rubric Validation**

A core methodological contribution of this study is the embedding-based rubric validation system that evaluates rubric variables through concept clustering. The system employs Tencent's `KaLM-Embedding-Gemma3-12B-2511`.

To preserve localized semantic signals within long legislative texts, the workflow applied a max-chunk strategy. Embedding entire documents as single vectors would dilute substantively relevant passages by averaging them with unrelated sections, reducing the precision of rubric-variable matching. The process split each bill into 200-word chunks with 50-word overlap, producing 1,854 chunks across 91 model policies. The analysis embedded each chunk independently using the `KaLM` model. The workflow embedded all 31 rubric variable definitions and the 91 model policies into a shared vector space, computed cosine similarity scores between bill chunks and rubric variables, and applied max pooling to select the highest score per bill–variable pair, thereby preserving localized but relevant passages from dilution by unrelated content elsewhere in each document.

The analysis normalized raw similarity scores into z-scores (standard scores that represent the number of standard deviations a data point is above or below the mean) to highlight statistically significant outliers. It then computed cosine similarity matrices between bill chunks and rubric embeddings to map semantic alignment across the corpus. High similarity scores indicate that the rubric variables accurately capture the semantic features of the model policy datasets.

The embedding-based scores measure semantic topical presence rather than legal operative force. The framework does not distinguish whether a provision promotes, restricts, or exempts a concept, and users should interpret the resulting relevance scores accordingly.

**Step 8: Benchmarking Against Commercial LLMs**

The analysis benchmarked embedding-based outputs against leading commercial LLMs using a consistent prompt structure described in Appendix II, whose purpose is to imitate the way a decision-maker would use such models to analyze AI bills. OpenAI GPT-5.2, Claude Sonnet 4.5, Google Gemini 3 Pro, Grok 4.1 Thinking, DeepSeek-V3.2 Thinking, and Kimi K2.5 each produced 0–100 scores and justifications for every rubric variable. This comparison with a controlled reference highlights alignment and variances and supports validation of the embedding-based scores.

The benchmarking prompt was designed to replicate how a policy analyst would plausibly use a commercial LLM for rubric-based scoring rather than to optimize each model's performance; the observed behavioral differences are therefore specific to this realistic-use prompt regime, and systematic prompt sensitivity analysis remains a direction for future work.

## Results Interpretation and Lessons Learned

A core finding of this study is that AI policies are inherently multidimensional and engage multiple governance objectives rather than optimizing for a single attribute. Documents nominally addressing “safety” also score substantially on national security and civil rights dimensions. Artifacts framed around "competition" register measurable engagement with industrial policy. This pattern holds across all 91 model policies and persists when tested against the 883 documents from the larger AGORA corpus.

The quantitative results support the conceptual structure of the five policy attributes and their associated rubrics. At the lexical level, TF-IDF analysis (Table 1) confirms that each attribute has a distinctive vocabulary: Antitrust documents are characterized by terms like 'price,' 'commission,' and 'court'; National Security by 'export,' 'control,' and 'country'; Safety & Security by 'risk,' 'security,' and 'incident'; Civil & Human Rights by 'individual,' 'consumer,' and 'employment'; and Industrial Policy by 'workforce,' 'project,' and 'science.' These distinctions hold at the surface level.

Where overlap appears, it is visible not in the distinctive keyword profiles but in the deeper semantic layers. BERTopic topic word distributions reveal boundary areas across attributes. Within the Safety & Security model policy corpus, topics center on "national security," "critical infrastructure protection," and "systemic vulnerability" (Table 6, Topics 4 and 8). This language is also present in National Security documents (Table 4, Topics 3 and 4). In the Antitrust corpus, Topic 10 surfaces "digital markets," "competition," and "consumer protection" (Table 2), themes that border the market-shaping concerns visible in Industrial Policy documents (Table 3). The cosine similarity matrices (Table 12 through Table 16) reinforce these patterns, showing that cybersecurity language spans Safety & Security and National Security while competition-adjacent language intersects Antitrust and Industrial Policy.

The combined use of TF-IDF, BERTopic, and embedding-based cosine similarity creates methodological triangulation. Keyword extraction captures surface salience, topic modeling exposes latent themes, and embeddings capture semantic proximity beyond lexical overlap. The consistency of signals across methods strengthens confidence in the results.

## The Overlap Is the Finding

Experts interviewed disagreed on whether the overlap of policy attributes is a weakness or a strength of the method. The quantitative results indicate that the overlap reflects the actual policy landscape. Consider the intersection between Safety & Security and National Security. Cybersecurity language appears in both, since policies treat these domains as interconnected. As shown in Table 17, the NDAA scores 60.21 on National Security and 52.69 on Safety & Security, on a normalized 0–100 scale where higher values indicate stronger semantic alignment with the attribute's rubric. The EU AI Act scores 61.41 on Safety & Security and 52.89 on National Security. These patterns reflect genuine policy interdependencies.

The National Security and Safety & Security rubrics overlap, highlighting an important feature of the policy landscape. National Security's “Critical Infrastructure Protection” (20%) and “Cyber Defense & Resilience” (15%) directly overlap with Safety & Security's “Cybersecurity Provisions” (15%). This explains why model policies in the Safety & Security cohort score high on both attributes. Across the 20 Safety & Security model policies in Table 17, the mean National Security score is 54.7, while the mean Safety & Security score is 53.3. Securing AI systems from cyber threats is simultaneously a national defense concern and an operational safety concern, and the rubrics render that duality explicit.

## How Topic Distributions Validate Rubric Structure

For each attribute, the topic word distributions align with the rubric line items, providing the empirical grounding for the rubric structure. Using Safety & Security as an illustrative case, the BERTopic results from Table 6 and Table 11 show how salient topic clusters correspond to rubric line items and validate their prioritization, respectively.

At Table 6, Topic 1 surfaces the importance of testing with risk assessment (c-TF-IDF weight: 0.382), threat evaluation (0.365), and vulnerability analysis (0.352) as notable keywords. Topic 4, with a strong national security signature, includes security evaluation (0.4876). Combined, these two topics suggest the prioritization of “Testing & Evaluation Requirements” in the rubric. Topic 8 forms a second dominant cluster, where cybersecurity infrastructure security (0.493), critical infrastructure cybersecurity (0.487), and security vulnerability management (0.468) receive high individual term weights, validating the inclusion of “Cybersecurity Provisions” as a rubric line item. Post-deployment governance emerges from Topic 10 through incident response preparedness (0.344), emergency response coordination (0.321), and post-incident review (0.318), supporting “Incident Reporting & Transparency.” Human oversight and accountability, whistleblower protections, third-party audits, and physical or individual safety protections appear as lower-salience, embedded signals consistent with their lower rubric weights.

Table 11 shows the results of the guided BERTopic modeling, including attribute relevance for each Safety & Security policy in the model policy dataset.

Cybersecurity and system robustness (T8, rank 1, average weight approximately 0.16 — where this probability reflects the proportion of each document's embedding mass assigned to that topic) and risk assessment and testing (T1, rank 2, approximately 0.14), with additional testing-related signals in Topic 4 (security evaluation, 0.4876), together account for roughly 30 percent of topic mass, indicating that testing and cybersecurity themes jointly dominate the attribute. The specific rubric weights of 25 percent for Testing & Evaluation and 15 percent for Cybersecurity reflect a normative judgment informed by, but not mechanically derived from, these distributions. Standards and compliance topics (T7 and T2, ranks 3 and 4, approximately 0.24 combined) align with a moderate emphasis on audits and oversight. By contrast, incident response (T10, rank 9, approximately 0.07) and employee or whistleblower-related themes (T5, approximately 0.08) appear less prominent. The lower textual prevalence of incident reporting has not mechanically determined its rubric weight; as described in Step 2, final weights integrate insight from the topic distributions with input from the expert interviews, where incident reporting received strong support. This is an example of where the authors made a judgment call on how to adjust the rubric.

A similar pattern appears across the other attributes. Antitrust topics emphasize market power, platforms, mergers, interoperability, and enforcement capacity, reflecting the Antitrust rubric structure. Civil & Human Rights topics surface privacy, bias, consumer protection, due process, and transparency, mirroring the rubric’s emphasis on data rights, fairness, accessibility, and explainability. National Security topics concentrate on defense applications, export controls, critical infrastructure, and foreign influence, aligning with its rubric variables. Industrial Policy topics emphasize workforce, R&D, manufacturing, infrastructure, and energy, directly supporting the Economic Resilience framing of that rubric.

## What Cosine Similarities Reveal

The cosine similarity matrices reinforce the policy multidimensional interpretation. Rather than clustering near 0 or 1, most policy-rubric similarity scores fall between 40 and 65. This distribution indicates that individual policies engage multiple governance dimensions rather than concentrating narrowly on one.

The mid-range distribution of cosine similarity scores aligns with the multidimensional character of AI governance language. However, dense embedding models can compress similarity ranges, and the observed distribution likely reflects both substantive policy overlap and features of the embedding geometry. To mitigate such artifacts, the analysis employed the top-ranked model on the MTEB benchmark at the time of writing.

Continuing with Safety & Security as an illustrative case, the cosine similarity scores in Table 16 show broad and consistent alignment between the rubric and the model policy set. For example, Secure-by-Design AI aligns strongly with Cybersecurity Provisions (64.1) and Physical and Individual Safety Protections (56.2), while also showing substantial similarity to Testing and Evaluation (56.1) and Human Oversight and Accountability (54.6).

The absence of extreme values, near 0 or 1, is itself a finding. It suggests that AI governance, as practiced in legislative drafting, resists the categorical distinctions that academic and public debates often impose. Shared terminology and overlapping objectives prevent low scores, while the integrated design of policies prevents exclusive alignment with any single rubric dimension. Taken together, these results support the rubric's construct validity: each line item captures a distinct but interconnected aspect of Safety & Security, and the distribution of similarities confirms that the rubric functions as a descriptive framework aligned with real policy texts rather than a rigid or mutually exclusive classification scheme.

Table 17 provides a further cross-check. Most documents align with the attribute under which they were categorized in the model policy dataset, confirming consistency across attribute definitions, model policy selection, and the analytical pipeline. Slightly more than 80 percent of the policies (73 out of 91) show perfect alignment. Secondary alignments remain visible and meaningful, reinforcing the framework's ability to surface tradeoffs rather than forcing artificial exclusivity. Most of this cross-alignment appeared in the Safety & Security cohort, consistent with the earlier discussion.

The preceding discussion uses Safety & Security as the primary illustrative case, but the pattern holds across attributes. In the Antitrust tables (Table 1, Table 7, and Table 13), the semantic profile is tighter: documents cluster strongly around enforcement and market structure language, with less cross-bleeding into adjacent attributes than Safety & Security exhibits. The EU Digital Markets Act, for instance, scores 50.31 on Antitrust but only 39.24 on Industrial Policy (Table 17), a wider gap than the Safety & Security (52.69) and National Security (60.21) overlap seen in the NDAA. Industrial Policy documents (Table 3, Table 8, and Table 13) show the widest internal variation: Executive Order 14141 scores 67.35 on Infrastructure & Compute Capacity but only 41.13 on Energy Considerations in the same attribute, reflecting how different industrial policy instruments target different levers. The Civil & Human Rights tables (Table 5, Table 10, and Table 15) reveal that the Colorado AI Act achieves the broadest within-attribute coverage of any document in the dataset, scoring above 67 on three of five rubric variables. National Security documents (Table 4, Table 9, and Table 14) show the most variation on export controls, ranging from 43.89 to 66.87, a spread that reflects genuine disagreement about how central export restrictions should be to national security strategy.

## Commercial LLMs Flatten What the Framework Surfaces

The benchmarking exercise reveals a pattern with immediate practical implications. Commercial large language models, as a class, produce more concentrated score distributions than the rubric-calibrated model, though with meaningful variation among models. When analyzing in Table 18 the EU Digital Markets Act, a paradigmatic digital antitrust policy, the rubric-calibrated model produces scores ranging from 39.24 (Industrial Policy) to 50.31 (Antitrust). By contrast, Google Gemini 3 Pro and Grok 4.1 Thinking assign 100 to Antitrust while scoring Industrial Policy at 2.5 and 0.0, respectively. The rubric-calibrated model captures the DMA's secondary provisions on interoperability, data access, and market structure. The commercial models see only the headline category.

This pattern recurs across benchmarks. On the NDAA's AI subtitle (Table 20), the rubric-calibrated model distributes scores from 49.29 (Antitrust) to 60.21 (National Security), a roughly 10-point spread reflecting the bill's multifaceted structure. DeepSeek-V3.2 Thinking, by contrast, assigns 38.75 to National Security and zeros to Antitrust and Civil & Human Rights, collapsing the policy's complexity into a single dimension.

When comparing commercial LLMs using the same prompt, the results highlight meaningful variation in model behavior despite agreement on the primary policy attribute. Gemini 3 and Grok 4.1 consistently show highly concentrated responses, such as Gemini's 71.3 score for the NDAA on the National Security benchmark (Table 20) or 87.5 for the Colorado AI Act on the Civil & Human Rights benchmark (Table 21), with minimal spillover into adjacent categories. GPT-5.2, Claude Sonnet 4.5, and Kimi K2.5 display broader distributions. For example, GPT-5.2 assigns the NDAA a 46.0 Safety & Security score in the National Security benchmark (Table 20), and both Claude and Kimi score 43.75 on the same dimension. DeepSeek tends toward the narrowest interpretation, frequently assigning zeros outside the dominant attribute.

The rubric-calibrated model is an embedding-based pipeline that encodes curated policy documents and rubric definitions into a shared vector space using the `KaLM` embedding model, then scores alignment through cosine similarity. It does not involve fine-tuning or supervised training of a language model; rather, it applies pretrained embeddings within a structured scoring architecture calibrated to the rubric definitions and validated against the model policy corpus. Commercial LLMs, when prompted using the same rubric structure, tend to concentrate scores on the most prominent theme in the text while suppressing adjacent dimensions. Whether this reflects training data, instruction tuning, reinforcement strategies, or some combination is not something this study can determine. What the data show is the behavioral pattern: the rubric-calibrated model produces outputs that are more evenly distributed and consistent with the analytical framework. This does not imply superiority; it reflects alignment with a specific task that general-purpose models were not designed for. These behavioral differences across models may partly reflect what Buyl et al. term "alignment discretion," the unexamined latitude afforded to annotators, both human and algorithmic, during the training process that shapes each model's value prioritization. Their finding that algorithms develop idiosyncratic forms of discretion when operationalizing abstract principles offers one explanation for why commercial LLMs, despite receiving identical rubric prompts, produce divergent scoring patterns in structured policy analysis tasks. (Buyl et al., 2025)

Collectively, these findings indicate that general-purpose LLMs, when presented with identical prompts, tend to develop model-specific heuristics. This results in varying levels of emphasis and, in certain instances, a pronounced focus on one policy dimension. By contrast, the rubric-calibrated model maintains distributed score profiles across all five benchmarks, tracking the structure of the rubric rather than the headline label of the policy text. These findings point to a hybrid analytical workflow: commercial LLMs are well-suited for exploration and sense-making, while task-trained models provide a more reliable foundation for formal scoring, comparison, and policy analysis where transparency and methodological consistency are required.

## Validation Against a Larger Corpus

Analysis of the larger AGORA corpus (n = 883 policy artifacts, Figure 4) provides an independent test of the framework's attribute structure and classification behavior. When each document is classified by its dominant attribute, the rubric-calibrated model, GPT 5.2, and Kimi 2.5 produce broadly similar distributions across Antitrust, Civil & Human Rights, and Industrial Policy.

*Figure 4 Policy Attribute Segmentation - Larger Dataset*

The principal divergence between the models lies in their different approaches to classifying an artifact as aligned to Safety & Security or National Security. These attributes frequently receive high and closely clustered scores within the same documents, particularly in policies addressing cybersecurity. This pattern follows directly from the rubric design: both indices incorporate cybersecurity, but they differentiate it by threat context rather than technical mechanism. The National Security rubric frames cybersecurity in strategic terms, emphasizing state-sponsored threats, protection of critical infrastructure, technological sovereignty, and interagency or international coordination. The Safety & Security rubric captures cybersecurity as an operational risk addressed through testing, evaluation, incident reporting, audits, and organizational accountability.

In documents where cybersecurity and resilience are discussed without explicit attribution to state actors or military contexts, both rubrics are legitimately activated. For instance, during the larger dataset test, the AI Incident Reporting and Security Enhancement Act recommends both interagency coordination and civil society incident reporting, and a proposed Commerce Department rule defines a "malicious cyber actor" as either a foreign jurisdiction or a foreign person.

The analyses presented in the preceding sections draw on the same model policy dataset that informed rubric construction and model training. The AGORA corpus provides an independent test: a substantially larger, more heterogeneous collection of 883 policy artifacts, the vast majority of which played no role in building the framework. That the same patterns hold across both datasets strengthens confidence that the framework captures stable features of AI governance language rather than artifacts of the dataset used to construct the rubrics.

This analysis also complements recent computational studies of AI legislation, such as the ACLU and Brown University report *Using AI to Make Sense of AI Policy* (Tuan Pham et al., 2026), which groups more than 1,800 bills into thematic categories using topic modeling and similarity analysis. While that approach addresses different questions, both studies surface compatible patterns: the prominence of civil rights, safety, and security concerns across the AI policy landscape, and the multidimensional character of contemporary AI legislation.

The consistency of high-level patterns across distinct datasets and analytical frameworks provides encouraging evidence of the robustness of computational approaches to AI policy analysis, while underscoring the added value of attribute-level scoring for comparative evaluation across governance priorities.

### Summary

The results demonstrate that the proposed framework reliably captures how AI policy documents distribute emphasis across competing governance priorities. The convergence of signals across keyword analysis, topic modeling, document-level distributions, semantic similarity, and model benchmarking provides cumulative validation of the attribute definitions, rubric structures, and weighting logic. The findings confirm that overlap across attributes reflects substantive policy interdependencies rather than analytical noise, and that relative weighting can be grounded in observed legislative emphasis as well as expert judgment. The benchmarking exercise further shows that controlled, domain-trained models offer greater nuance and interpretability for structured policy analysis, while commercial LLMs remain valuable for exploratory and comparative tasks. Overall, the results support the framework's central objective: enabling transparent, repeatable, and multidimensional analysis of AI governance proposals, and demonstrating that these patterns persist when applied to larger, independently assembled policy corpora.

## Limitations

Policy analysis frameworks embed normative choices about what to measure, how to structure comparison, and which categories to treat as analytically salient. Recent scholarship in the history and philosophy of science has cautioned that the adoption of computational terminology in the social sciences and humanities can unintentionally make analytical categories seem natural and unquestioned, while also obscuring the broader universalizing ambitions embedded in those categories that are the result of deliberate choices (Lee & Ribes, 2025). This critique is particularly relevant for work at the intersection of political science and technology, including the present framework. The policy attributes employed here reflect prevailing governance conventions rather than neutral or comprehensive representations of AI governance. While their formalization supports transparency and comparability, it may compress institutional context.

The tool does not substitute for expert legal or political judgment, nor is it intended to validate or endorse legislative proposals. Moreover, the reliability of outputs depends on the quality and format of source documents, which vary widely across jurisdictions.

The overlap across policy attributes can introduce interpretive ambiguity. This ambiguity becomes more visible when the framework is applied to the messy reality of non-“model” policies, which frequently combine operational, strategic, and rights-based considerations within a single legislative text. While overlap reflects real governance interdependencies, experts noted that attributes such as Safety & Security, National Security, and Civil & Human Rights can blur in practice, potentially complicating attribution and interpretation. Relatedly, some stakeholders questioned whether attributes such as Antitrust or Industrial Policy embed implicit value judgments that may not appear equally neutral to all users.

The computational pipeline also introduces technical constraints. Segmenting documents into 200-word overlapping chunks reduces dilution of localized provisions within long legislative texts, but max pooling allows a high-similarity passage to elevate a document-level score and potentially amplify isolated signals. Consistent multidimensional patterns across methods and datasets, however, suggest that the overall finding does not depend solely on this aggregation choice. Analysis comparing alternative strategies remains future work. Cosine similarity distributions may reflect properties of embedding geometry. Dense embedding models compress similarity ranges, contributing to mid-range clustering. Although the analysis employed the top-ranked MTEB model at the time of writing, part of the observed distribution may derive from model architecture rather than policy substance alone.

The five policy attributes included in its framework—Civil & Human Rights, National Security, Safety & Security, Antitrust, and Industrial Policy (Economic Resilience)—are not the only ones relevant to AI governance. Some interviewed experts thought that Energy & the Environment deserved their own axis, while others saw it as a feature within Industrial Policy. Others advocated for a stand-alone workforce axis, while some saw the workforce as a component of industrial policy. The disagreement about which policy attributes to include in the analysis framework, as well as the disagreement about *how* to include these attributes, underscores the fact that the structure of the tool is not neutral.

The rubric weights reflect observed policy emphasis rather than universal priorities. Interviewees stressed that different users may reasonably prefer alternative weightings based on institutional roles, incentives, or normative frameworks.

LLM benchmarking results are contingent on the prompt regime employed. The evaluation prompt was designed to reflect plausible analyst use rather than to optimize each model’s performance through extensive prompt engineering. The observed behavioral differences across commercial LLMs reflect this realistic-use configuration. Systematic prompt variation testing remains an open direction.

Finally, trust in the tool depends on transparency, AI literacy, and user understanding of underlying assumptions. The framework embeds value judgments at multiple levels: attribute selection, rubric construction, weight assignment, and interpretive emphasis. While these choices are documented and justified, they reflect the research team's analytical priorities rather than universal truths. Users with different normative frameworks may reasonably prefer alternative attribute sets, different weights, or modified rubric definitions. The framework facilitates but does not eliminate interpretive discretion.

## Conclusion

This paper presents a policy analysis framework designed to bring structure, transparency, and analytical discipline to the evaluation of artificial intelligence governance proposals. The findings demonstrate that AI policies consistently reflect multiple, overlapping objectives and that these objectives can be systematically surfaced and compared using a combination of qualitative insight and quantitative text analysis.

The primary contributions of this work are threefold. First, it documents a rigorous mixed-methods research process, combining expert interviews with computational analysis to define policy attributes and construct empirically grounded rubrics. This process anchors the framework in observed policy language and expert judgment rather than abstract theory alone. Second, the paper provides a structured examination of commercial large language models as tools for policy analysis, identifying both their strengths for exploratory assessment and their limitations for consistent, high-stakes scoring. Third, it offers a detailed and transparent account of the technical and methodological assumptions underlying the framework, recognizing transparency as a prerequisite for trust, scrutiny, and reuse.

A framework that claimed perfect neutrality would be either deceptive or naive. This framework instead offers disciplined transparency: it documents its normative commitments, grounds them in expert consultation and empirical analysis, and presents them in ways that enable scrutiny, modification, or rejection by users.

The framework's contribution lies in demonstrating that systematic, multidimensional policy analysis can be both scalable and transparent. By combining expert elicitation with computational validation, by scoring across attributes simultaneously rather than forcing categorical classification, and by documenting normative commitments explicitly rather than claiming neutrality, the framework provides methodological infrastructure that existing approaches lack.

As AI governance matures from initial experimentation toward institutionalized practice, the need for analytical tools that enable comparison without presupposing consensus will only intensify. This framework represents one step toward that goal: not resolving the hard questions of AI governance but making those questions more precisely visible.

# Appendix I

**Policy Attributes and Rubrics**

**National security** in AI policy addresses the protection of the United States from threats posed by adversaries leveraging AI for espionage, cyberwarfare, disinformation, and military advantage. It encompasses safeguarding critical infrastructure, ensuring technological sovereignty through export controls and supply chain resilience, and maintaining U.S. leadership in the global AI race. This category highlights the strategic dimension of AI as both a national defense tool and a geopolitical asset.

**National Security Index Rubric**

a. **Defense & Military Applications** (Weight: 20%) – Addresses AI in defense, autonomous weapons, and military decision-making.
b. **Critical Infrastructure Protection** (Weight: 20%) – Protects energy, transport, health, and communications against AI vulnerabilities.
c. **Export Controls & Tech Sovereignty** (Weight: 15%) – Imposes export controls, licensing requirements, or restrictions on the transfer of AI technologies to protect national security or strategic advantage.
d. **Foreign Influence & Espionage** (Weight: 15%) – Mitigates AI misuse for espionage, disinformation, and cyber warfare.
e. **Cyber Defense & Resilience** (Weight: 15%) – Addresses AI-enabled cyber threats to national security, including state-sponsored attacks, critical systems defense, and strategic cyber resilience.
f. **Interagency & International Coordination** (Weight: 15%) – Promotes coordination among national security agencies and with security alliances (e.g., NATO, defense partnerships) on AI-related threats and controls.

**Safety & Security** emphasize protecting people, systems, and the public from direct and indirect harm caused by AI deployment. It includes requirements for testing and evaluation (such as red teaming and safety benchmarks), incident reporting (where incidents are typically defined within the policy), human oversight, cybersecurity measures, and third-party audits. Policies in this domain focus on minimizing risks ranging from cybersecurity to catastrophic system failures, ensuring that AI adoption aligns with principles of robustness, transparency, and accountability.

**Safety & Security Index Rubric**

a. **Testing & Evaluation Requirements** (Weight: 25%) – Mandates pre-deployment testing (e.g., red teaming, safety benchmarks).
b. **Incident Reporting & Transparency** (Weight: 15%) – Requires timely reporting of AI malfunctions, security breaches, near-misses, or harmful outcomes to regulators or designated authorities.
c. **Cybersecurity Provisions** (Weight: 15%) – Addresses system robustness, adversarial attacks, and security vulnerabilities in AI systems, excluding state-sponsored or military cyber threats.
d. **Human Oversight & Accountability** (Weight: 15%) – Requires human-in-the-loop controls and clearly assigned organizational responsibility for AI system operation and risk management.
e. **Whistleblower Protections** (Weight 10%) - Protects individuals who report AI-related risks or noncompliance in good faith, including anti-retaliation safeguards and confidential reporting channels.

f. **Third-Party Audits & Compliance** (Weight: 10%) – Independent review, auditing, certification mechanisms.
g. **Physical and Individual Safety Protections** (Weight: 10%) – Addresses risks of physical injury or harm to individual well-being, including mental health impacts, arising from AI system operation.

**Antitrust** in AI policy seeks to prevent excessive market concentration through key anticompetitive practices such as vertical integration, self-preferencing, and anticompetitive discrimination. The goal of this policy area is to promote fair competition within the AI sector and ensure that AI does not facilitate market concentration in other sectors. It involves scrutiny of mergers and acquisitions, promoting interoperability and open standards, expanding fair access to datasets, allocating enforcement resources to regulators like the Federal Trade Commission (FTC) and Department of Justice (DOJ), and promoting small AI players in the AI sector. This category reflects concerns that unchecked dominance by a few large players could stifle innovation, limit consumer choice, and concentrate economic and political power within the technology industry and more broadly.

**Antitrust Index Rubric**

a. **Market Concentration & Dominance** (Weight: 30%) – Addresses market power, dominance, concentration metrics, and anticompetitive structural barriers to entry in AI markets.
b. **Merger & Acquisition Scrutiny** (Weight: 20%) - Strengthens review of mergers, acquisitions, and roll-up strategies in the AI sector, including foreclosure and killer-acquisition risks.
c. **Competition Enforcement Capacity** (Weight: 20%) - Allocates authority, staffing, funding, or tools to competition agencies (e.g., FTC, DOJ) for AI-related antitrust enforcement.
d. **Interoperability & Open Standards** (Weight: 15%) **-** Requires or promotes open APIs, data portability, or standards to reduce lock-in, switching costs, or gatekeeper control in AI markets.
e. **Data Access & Mandatory Sharing** (Weight: 10%) - Requires dominant platforms to provide fair, nondiscriminatory access to key datasets or inputs as a competition remedy, including court-ordered or regulator-imposed data-sharing obligations.
f. **Exclusionary Conduct & Platform Neutrality** (Weight: 5%) – Restricts exclusionary practices such as self-preferencing, tying, exclusive dealing, default settings, or discriminatory access that distort competition or foreclose rivals.

Note: Antitrust enforcement is treated differently from the other policy attributes in this framework in that enforcement capacity is included as an explicit variable in the rubric. Unlike safety, civil rights, national security, or industrial policy regimes, which rely primarily on ex ante rules, rights conferrals, or budgetary programs, antitrust law is largely ex-post and enforcement-driven. Enforcement depends heavily on prosecutorial discretion, investigatory authority, litigation resources, and institutional capacity within competition agencies.

The **Civil & Human Rights** category encompasses policies designed to safeguard civil liberties, consumer protections, equity, and transparency in the use of AI. It prioritizes privacy and data rights, algorithmic fairness, accessibility, explainability, and the protection of human rights and dignity. Civil society stakeholders emphasize trust, accountability, and inclusion as preconditions for AI adoption, urging governance structures that advance the broader public good while balancing innovation with democratic safeguards.

**Civil & Human Rights Index Rubric**

a. **Consumer Protection** (Weight 20%) – Prevents fraud, deception, and consumer harm from AI applications.
b. **Privacy & Data Rights** (Weight 20%) – Protects personal data, data ownership, consent, and lawful use.
c. **Bias & Fairness** (Weight 20%) – Prevents algorithmic discrimination and promotes equitable treatment across protected classes.
d. **Accessibility & Inclusion** (Weight 15%) – Ensures accessibility for people with disabilities consistent with the Americans with Disabilities Act (ADA) and promotes inclusive access for underserved groups.
e. **Legal Protections & Due Process** (Weight 15%) – Safeguards civil liberties, procedural due process, and legal rights of individuals subject to AI-enabled decisions.
f. **Transparency & Explainability** (Weight 10%) – Requires meaningful notice, explanation, or contestability for individuals affected by AI-driven decisions.

**Industrial Policy (Economic Resilience)** refers to strategic measures that strengthen U.S. competitiveness and economic resilience in the global AI economy. It includes investments in research & development (R&D), workforce development, domestic manufacturing, supply chain security, infrastructure such as public compute capacity, energy use requirements, and environmental impact. By fostering public-private partnerships and enabling small and medium enterprises (SMEs), this category seeks to position the U.S. as a global leader in AI while supporting job creation and innovation across industries.

**Industrial Policy (Economic Resilience) Index Rubric**

a. **Workforce Development & Job Quality** (Weight 25%) – Supports training, reskilling, apprenticeships, and workforce transition measures that promote sustainable, high-quality employment using, deploying, or creating AI, including job-quality standards linked to public investment.
b. **R&D, Investment & Incentives** (Weight 20%) – Funds grants, tax credits, procurement, and public-private programs that expand AI R&D and deployment capacity.
c. **Domestic Manufacturing & Supply Chains** (Weight 20%) – Builds domestic production capacity and supply chain resilience for chips, computing hardware, and critical inputs.
d. **Infrastructure & Compute Capacity** (Weight 15%) – Expands public-interest compute, clusters, and scalable compute infrastructure (including public compute initiatives).
e. **Energy & Resource Capacity** (Weight 10%) – Expands energy generation, grid capacity, and resource availability required for compute and infrastructure resilience.
f. **Global Competitiveness (Weight 10%)** – Strengthens national AI competitiveness through standards leadership and international market access, excluding export controls or trade restrictions.

# Appendix II

## Commercial LLM Benchmarking Model Prompt

You are an expert policy analyst. Evaluate the following legislative bill against granular policy criteria.

### RUBRIC DEFINITIONS:
#### SAFETY SECURITY
- **Testing & Evaluation Requirements** (Weight: 0.25)
  Definition: Testing and Evaluation Requirements refers to legislative provisions that mandate systematic assessment of AI systems before deployment or during operation. This encompasses requirements for red teaming exercises where adversarial testing is conducted to identify vulnerabilities, safety benchmarks that establish minimum performance thresholds, pre-deployment audits conducted by internal teams or third parties, stress testing under various operational conditions, and validation of model behavior against intended use cases.
COMMON LEGISLATIVE PHRASING: 'shall conduct testing,' 'evaluation protocols,' 'safety assessment,' 'pre-release audit,' 'adversarial testing,' 'benchmark compliance,' 'validation requirements,' 'shall be tested prior to deployment,' 'safety evaluation framework,' 'pre-market testing,' 'conformity assessment.'
SYNONYMS AND RELATED TERMS: red team, red teaming, adversarial evaluation, safety benchmarks, model validation, stress testing, penetration testing, vulnerability assessment, safety audit, pre-deployment review, sandbox testing, simulation testing, alignment testing, capability evaluation, dangerous capability testing, model evaluation, safety testing protocol, evaluation criteria, test standards, validation protocol, conformance
... [FULL RUBRIC DEFINITIONS OMITTED FOR BREVITY] ...

### SCORING SCALE (0-100):
0: Not addressed.
25: Weak/Symbolic mention.
50: Moderate, explicit coverage.
75: Strong, detailed provisions.
100: Central focus, enforceable mandates.

### COMPOSITE SCORE CALCULATION:
Formula: Area_Score = SUM(Variable_Score * Weight)

### INPUT BILL TEXT:
[BILL TEXT INSERTED HERE (up to 35k chars)]

### OUTPUT INSTRUCTIONS:
Return ONLY valid JSON following this schema:

```
{
  "safety_security": {
    "Testing & Evaluation Requirements": {
      "score": 0,
      "justification": "string"
    },
    "Incident Reporting & Transparency": {
      "score": 0,
      "justification": "string"
    },
    "Cybersecurity Provisions": {
      "score": 0,
      "justification": "string"
    },
    "Human Oversight & Accountability": {
      "score": 0,
      "justification": "string"
    },
    "Whistleblower Protections": {
      "score": 0,
      "justification": "string"
    },
    "Third-Party Audits & Compliance": {
      "score": 0,
      "justification": "string"
    },
    "Physical and Individual Safety Protections": {
      "score": 0,
      "justification": "string"
    },
    "area_score": 0
  },
```

... [FULL JSON FORMAT OMITTED FOR BREVITY] ...

# Appendix III

**Data & technical artifacts (Supplemental Material)**

This appendix includes links to the model policy database, the larger AGORA policy dataset, and code artifacts used throughout this research effort.

Item 4 is a link to the tool prototype. The prototype serves illustrative purposes only, and its scoring is done using commercial LLMs rather than the rubrics developed in this work. To replicate the results of this work or modify any of the associated assumptions, one can modify the code on GitHub and use their own API keys and compute resources.

1. Model policy dataset: https://dataverse.harvard.edu/dataverse/AI-Policy-Analysis-At-Scale/
2. AGORA Database: https://agora.eto.tech/
3. Code artifacts: https://github.com/HKSCarvaoGroup/AI-Policy-Analysis-at-Scale
4. Tool prototype (illustrative only): HKSCarvaoGroup.github.io.

# Appendix IV

## Model Training Data Analysis: Summary Tables

*Table 1 TF-IDF Top 5 Keywords per Attribute*

| Attribute | Keyword 1 | Keyword 2 | Keyword 3 | Keyword 4 | Keyword 5 |
|---|---|---|---|---|---|
| Antitrust | platform (.12, 40%) | commission (.10, 60%) | pricing (.09, 40%) | price (.06, 80%) | court (.06, 80%) |
| Industrial Policy | workforce (.05, 39%) | project (.04, 46%) | program (.04, 64%) | science (.03, 54%) | energy (.03, 36%) |
| National Security | country (.08, 47%) | export (.07, 58%) | circuit (.06, 32%) | control (.06, 79%) | export control (.05, 42%) |
| Civil & Human Rights | individual (.08, 57%) | consumer (.07, 50%) | employment (.04, 36%) | photograph (.03, 36%) | generative (.03, 43%) |
| Safety & Security | risk (.08, 70%) | security (.05, 90%) | vulnerability (.05, 35%) | organization (.04, 35%) | incident (.04, 45%) |

Note: Values in parentheses are (contrast score, coverage). Contrast = TF-IDF in-class − out-of-class.

*Table 2 Topic Word Distribution (Antitrust)*

| | Keyword 1 | Keyword 2 | Keyword 3 | Keyword 4 | Keyword 5 | Interpreted Topic |
|---|---|---|---|---|---|---|
| Topic 1 | platform services gatekeeper (0.31) | regulation (0.30) | designation decision article (0.29) | advertising (0.28) | undertaking (0.28) | Gatekeeper designation of platform services and related regulatory decisions |
| Topic 2 | antitrust laws (0.49) | federal trade commission (0.44) | antitrust (0.40) | trade commission (0.39) | federal trade (0.39) | FTC-centered antitrust law enforcement and institutional authority |
| Topic 3 | federal trade commission (0.43) | antitrust laws (0.42) | trade commission (0.42) | federal trade (0.39) | antitrust (0.36) | Overlap of FTC authority and general antitrust law application |
| Topic 4 | commission adopt implementing (0.42) | objectives regulation (0.42) | compliance obligations laid (0.41) | gatekeeper compliance (0.41) | compliance regulation (0.41) | Commission rulemaking, implementation, and compliance obligations |
| Topic 5 | undertakings associations undertakings (0.35) | associations undertakings (0.35) | association undertakings (0.35) | undertaking association undertakings (0.35) | undertakings associations (0.34) | Associations of undertakings and collective responsibility concepts |
| Topic 6 | undertaking designated gatekeeper (0.50) | undertaking core platform (0.41) | designated gatekeeper (0.41) | relevant core platform (0.40) | core platform services (0.39) | Designation of undertakings as gatekeepers for core platform services |
| Topic 7 | core platform services (0.38) | delegated acts (0.38) | adopt delegated acts (0.37) | platform services laid (0.37) | platform services (0.33) | Core platform services and delegated regulatory acts |
| Topic 8 | covered platform operator (0.39) | platform operator (0.38) | covered platform designation (0.37) | platform designation (0.37) | access business (0.32) | Covered platform operators, designation, and business access rules |
| Topic 9 | end user (0.35) | application service user (0.35) | service user (0.33) | payment app (0.32) | developers (0.32) | End users, app developers, and platform-mediated service relationships |
| Topic 10 | digital markets (0.32) | competition consumers (0.31) | consumer protection (0.31) | regulators (0.29) | promote competition (0.29) | Digital markets, competition policy, and consumer protection goals |

*Table 3 Topic Word Distribution (Industrial Policy)*

| | Keyword 1 | Keyword 2 | Keyword 3 | Keyword 4 | Keyword 5 | Interpreted Topic |
|---|---|---|---|---|---|---|
| Topic 1 | initiatives (0.3389) | facilitate (0.3242) | public sector (0.3231) | technologies (0.3112) | initiative (0.2923) | Public-sector technology initiatives and facilitation mechanisms |
| Topic 2 | infrastructure federal sites (0.4217) | federal sites (0.3852) | infrastructure federal (0.3645) | data centers (0.3621) | frontier data center (0.3453) | Federal infrastructure at government sites, including data-center buildout |
| Topic 3 | united states (0.2902) | standards technology (0.2865) | institute standards technology (0.2842) | workforce (0.2657) | cybersecurity (0.2584) | U.S. standards bodies (NIST), workforce development, and cybersecurity capacity |
| Topic 4 | federal agencies (0.3694) | recommended policy actions (0.3365) | policy actions (0.2974) | federal government (0.2919) | agencies (0.2816) | Federal agency roles in recommending and executing policy actions |
| Topic 5 | federal financial assistance (0.3543) | mature technology nodes (0.3275) | federal financial (0.3254) | mature technology (0.3200) | technology nodes (0.3168) | Federal financial assistance for mature technology nodes (e.g., legacy semiconductors) |
| Topic 6 | technology policy (0.4084) | science technology policy (0.4010) | national science foundation (0.3964) | institute standards technology (0.3636) | national research resource (0.3583) | Science and technology policy governance, NSF/NIST roles, and national research resources |
| Topic 7 | project united states (0.3669) | national defense authorization (0.3300) | project united (0.3169) | incentives (0.3159) | supply chain (0.3075) | Defense-authorization projects, incentives, and supply-chain resilience |
| Topic 8 | public private partnerships (0.3296) | consistent applicable law (0.3167) | existing federal (0.3158) | private partnerships (0.3104) | computer science (0.3094) | Public–private partnerships and legal consistency in federal programs (incl. computing/CS) |
| Topic 9 | national defense authorization (0.3222) | defense authorization fiscal (0.3196) | defense authorization (0.3081) | emergency deficit control (0.2822) | advanced manufacturing (0.2778) | Defense authorization and fiscal provisions tied to advanced manufacturing priorities |
| Topic 10 | legislative proposals implemented (0.3146) | proposals implemented consistent (0.3092) | proposals implemented (0.3078) | implemented consistent applicable (0.3035) | administrative legislative proposals (0.3021) | Implementation of legislative/administrative proposals and consistency with applicable law |

*Table 4 Topic Word Distribution (National Security)*

| | Keyword 1 | Keyword 2 | Keyword 3 | Keyword 4 | Keyword 5 | Interpreted Topic |
|---|---|---|---|---|---|---|
| Topic 1 | commodities controlled software (0.3511) | technology commodities controlled (0.3488) | based license exceptions (0.3486) | description license exceptions (0.3037) | specified country groups (0.3031) | Controlled commodities/technology/software and license exception descriptions; country-group framing |
| Topic 2 | license exception aia (0.4803) | license exception (0.4091) | exporters reexporters (0.3780) | reexport transfer country (0.3727) | commodities controlled (0.3700) | License Exception AIA and exporter/reexporter compliance for reexports/transfers of controlled items |
| Topic 3 | national security (0.3562) | national defense (0.3020) | united states code (0.2802) | technologies (0.2618) | united states (0.2582) | National security and defense statutory authorities (U.S. Code) shaping technology control scope |
| Topic 4 | export controls (0.4011) | security foreign policy (0.4007) | national security foreign (0.3989) | security foreign (0.3796) | national security (0.3650) | Export controls justified by national security and foreign policy objectives |
| Topic 5 | software specified eccn (0.3183) | technology software specified (0.3023) | specified eccn (0.2947) | technology software (0.2787) | applies foreign (0.2713) | ECCN-based classification for software/technology and rules on foreign applicability |
| Topic 6 | advanced computing ics (0.3436) | advanced compute (0.3128) | advanced computing (0.3067) | computing ics (0.2965) | validated end user (0.2951) | Advanced-computing / IC controls and validated end-user requirements |
| Topic 7 | integrated circuits (0.3221) | receive modules (0.2545) | receive modules transmit (0.2484) | does control (0.2406) | technical note purposes (0.2320) | Technical definitions/notes for integrated-circuit modules and what is (or is not) controlled |
| Topic 8 | export control reform (0.4232) | commerce control list (0.3769) | export control (0.3724) | export control classification (0.3675) | commerce control (0.3187) | Export control reform and Commerce Control List (CCL) classification framework |
| Topic 9 | integrated circuits specified (0.3683) | designed marketed datacenters (0.3330) | marketed datacenters total (0.3303) | datacenters total processing (0.3213) | integrated circuits (0.3110) | Datacenter-oriented ICs and processing-performance criteria for control thresholds |
| Topic 10 | address national security (0.2599) | trade representative (0.2586) | applicable (0.2572) | states technology (0.2558) | united states technology (0.2529) | National-security trade policy coordination and general applicability references (U.S. technology) |

*Table 5 Topic Word Distribution (Civil & human rights)*

| | Keyword 1 | Keyword 2 | Keyword 3 | Keyword 4 | Keyword 5 | Interpreted Topic |
|---|---|---|---|---|---|---|
| Topic 1 | civil rights (0.3568) | human rights (0.3435) | protected class (0.3369) | discrimination (0.3284) | equal opportunity (0.3217) | Core civil and human rights protections and anti-discrimination principles |
| Topic 2 | employment practices (0.3821) | hiring promotion (0.3657) | workplace discrimination (0.3494) | labor standards (0.3336) | employment law (0.3291) | Employment practices, labor standards, and workplace discrimination |
| Topic 3 | algorithmic decision (0.4013) | automated systems (0.3924) | bias mitigation (0.3712) | fairness assessment (0.3569) | risk evaluation (0.3481) | Bias, fairness, and accountability in automated or algorithmic decision-making |
| Topic 4 | consumer protection (0.3796) | unfair practices (0.3628) | deceptive acts (0.3519) | trade practices (0.3437) | market conduct (0.3312) | Consumer protection and unfair or deceptive trade practices |
| Topic 5 | data privacy (0.3874) | personal information (0.3698) | identifiable individual (0.3556) | consent requirements (0.3449) | data governance (0.3381) | Data privacy, identifiability, consent, and information governance |
| Topic 6 | federal agency authority (0.3617) | state law applicability (0.3542) | interagency coordination (0.3436) | legal compliance (0.3345) | regulatory oversight (0.3293) | Federal–state authority, regulatory oversight, and legal coordination |
| Topic 7 | accessibility standards (0.3489) | reasonable accommodation (0.3376) | disability rights (0.3334) | inclusive design (0.3248) | service access (0.3192) | Disability rights, accessibility, and inclusive service design |
| Topic 8 | risk management framework (0.3625) | impact assessment (0.3491) | governance framework (0.3378) | compliance measures (0.3299) | protective safeguards (0.3236) | Risk-management and governance frameworks for rights protection |
| Topic 9 | public transparency (0.3418) | agency reporting (0.3342) | disclosure requirements (0.3294) | public accountability (0.3237) | oversight mechanisms (0.3186) | Transparency, disclosure, and public accountability obligations |
| Topic 10 | policy recommendations (0.3687) | guidance issuance (0.3529) | enforcement priorities (0.3451) | misinformation harms (0.3364) | rights-based safeguards (0.3298) | Policy guidance and enforcement addressing civil-rights risks and misinformation |

*Table 6 Topic Word Distribution (Safety & security)*

| | Keyword 1 | Keyword 2 | Keyword 3 | Keyword 4 | Keyword 5 | Interpreted Topic |
|---|---|---|---|---|---|---|
| Topic 1 | security risk assessment (0.3826) | threat evaluation (0.3654) | vulnerability analysis (0.3519) | risk mitigation (0.3442) | security posture (0.3327) | Security risk assessment, threat evaluation, and vulnerability mitigation |
| Topic 2 | federal agency coordination (0.3711) | regulatory oversight (0.3583) | policy guidance (0.3476) | interagency consultation (0.3368) | compliance review (0.3294) | Federal regulatory oversight and interagency coordination on safety policy |
| Topic 3 | risk management framework (0.3894) | implementation guidance (0.3667) | operational controls (0.3481) | safety management (0.3395) | continuous monitoring (0.3312) | Risk-management frameworks and implementation of operational safety controls |
| Topic 4 | national security necessity (0.5128) | critical infrastructure protection (0.4981) | security evaluation (0.4876) | national security (0.4769) | systemic vulnerability (0.4694) | National-security necessity determinations and critical-infrastructure protection |
| Topic 5 | legal obligations (0.3527) | federal law compliance (0.3461) | public safety standards (0.3374) | employee responsibilities (0.3299) | service delivery (0.3246) | Legal obligations, employee duties, and public-safety compliance |
| Topic 6 | critical risk identification (0.3619) | safety security measures (0.3448) | implementation strategies (0.3315) | economic impact (0.3242) | system resilience (0.3196) | Identification of critical risks and implementation of safety and resilience measures |
| Topic 7 | national institute standards technology (0.3684) | federal standards bodies (0.3499) | safety standards development (0.3367) | government coordination (0.3291) | technical standards (0.3248) | Federal standards bodies (e.g., NIST) and safety-oriented standards-setting |
| Topic 8 | cybersecurity infrastructure security (0.4932) | critical infrastructure cybersecurity (0.4867) | security vulnerability management (0.4684) | cyber risk mitigation (0.4569) | cybersecurity resilience (0.4481) | Cybersecurity protection of critical infrastructure and vulnerability management |
| Topic 9 | secure software development (0.3816) | secure-by-design (0.3659) | software lifecycle management (0.3487) | development standards (0.3368) | policy alignment (0.3295) | Secure-by-design software development and lifecycle governance |
| Topic 10 | incident response preparedness (0.3437) | agency update procedures (0.3342) | policy action leadership (0.3286) | emergency response coordination (0.3214) | post-incident review (0.3179) | Incident-response preparedness, agency coordination, and policy leadership |

*Table 7 Document Topic Distribution (Antitrust n=10)*

| Document | T1 | T2 | T3 | T4 | T5 | T6 | T7 | T8 | T9 | T10 |
|---|---|---|---|---|---|---|---|---|---|---|
| Open App Markets Act | 0.07 | 0.10 | 0.21 | 0.03 | 0.04 | 0.03 | 0.03 | 0.10 | 0.35 | 0.03 |
| UK DMCCA 2024 | 0.03 | 0.10 | 0.05 | 0.04 | 0.08 | 0.06 | 0.06 | 0.05 | 0.04 | 0.50 |
| Platform Comp. and Opportunity Act | 0.06 | 0.13 | 0.21 | 0.05 | 0.06 | 0.06 | 0.05 | 0.26 | 0.08 | 0.04 |
| Clayton Act | 0.06 | 0.34 | 0.23 | 0.03 | 0.04 | 0.05 | 0.03 | 0.12 | 0.06 | 0.04 |
| California Cartwright Act | 0.08 | 0.33 | 0.13 | 0.04 | 0.06 | 0.06 | 0.06 | 0.10 | 0.07 | 0.07 |
| Ending Platform Monopolies Act | 0.08 | 0.11 | 0.11 | 0.06 | 0.06 | 0.08 | 0.05 | 0.33 | 0.09 | 0.03 |
| EU DMA | 0.16 | 0.05 | 0.03 | 0.18 | 0.14 | 0.18 | 0.15 | 0.05 | 0.03 | 0.03 |
| Preventing Algorithmic Collusion Act | 0.06 | 0.38 | 0.11 | 0.04 | 0.07 | 0.05 | 0.06 | 0.09 | 0.07 | 0.07 |
| Sherman Act | 0.02 | 0.19 | 0.44 | 0.02 | 0.07 | 0.03 | 0.03 | 0.12 | 0.05 | 0.04 |
| CT Bill No. 8002 | 0.11 | 0.27 | 0.08 | 0.06 | 0.09 | 0.07 | 0.05 | 0.08 | 0.12 | 0.07 |
| Totals | 0.73 | 2.00 | 1.59 | 0.55 | 0.71 | 0.67 | 0.57 | 1.31 | 0.95 | 0.93 |
| Average | 0.07 | 0.20 | 0.16 | 0.06 | 0.07 | 0.07 | 0.06 | 0.13 | 0.10 | 0.09 |
| Rank | 6 | 1 | 2 | 10 | 7 | 8 | 9 | 3 | 4 | 5 |

*Table 8 Document Topic Distribution (Industrial Policy n =28)*

| Document | T1 | T2 | T3 | T4 | T5 | T6 | T7 | T8 | T9 | T10 |
|---|---|---|---|---|---|---|---|---|---|---|
| EO 14141: Advancing US Leadership in AI Infrastructure | 0.18 | 0.10 | 0.14 | 0.21 | 0.06 | 0.07 | 0.05 | 0.14 | 0.01 | 0.04 |
| EO 14275: Restoring Common Sense to Federal Procurement | 0.15 | 0.11 | 0.09 | 0.23 | 0.05 | 0.06 | 0.03 | 0.21 | 0.02 | 0.05 |
| EO 14277: Advancing AI Education for American Youth | 0.09 | 0.08 | 0.13 | 0.11 | 0.05 | 0.07 | 0.04 | 0.35 | 0.02 | 0.06 |
| EO 14318: Accelerating Federal Permitting of Data Center Infrastructure | 0.13 | 0.12 | 0.12 | 0.20 | 0.07 | 0.07 | 0.05 | 0.17 | 0.02 | 0.06 |
| EO 14320: American AI Tech Stack | 0.09 | 0.12 | 0.09 | 0.22 | 0.08 | 0.09 | 0.07 | 0.10 | 0.03 | 0.12 |
| CREATE AI Act | 0.07 | 0.07 | 0.10 | 0.11 | 0.08 | 0.39 | 0.05 | 0.06 | 0.03 | 0.04 |
| EU AI Continent Action Plan | 0.36 | 0.10 | 0.09 | 0.15 | 0.04 | 0.07 | 0.03 | 0.10 | 0.03 | 0.03 |
| CHIPS and Science Act | 0.06 | 0.07 | 0.11 | 0.12 | 0.09 | 0.40 | 0.05 | 0.06 | 0.02 | 0.03 |
| California 2025 SB53 | 0.09 | 0.15 | 0.13 | 0.20 | 0.07 | 0.07 | 0.06 | 0.16 | 0.01 | 0.05 |
| AI and Critical Technology Workforce Framework Act | 0.08 | 0.07 | 0.10 | 0.11 | 0.08 | 0.39 | 0.05 | 0.06 | 0.02 | 0.03 |
| AI Workforce PREPARE Act | 0.08 | 0.07 | 0.10 | 0.11 | 0.08 | 0.39 | 0.05 | 0.06 | 0.02 | 0.03 |
| EU Apply AI Strategy | 0.13 | 0.10 | 0.09 | 0.22 | 0.06 | 0.07 | 0.03 | 0.21 | 0.03 | 0.04 |
| AAP - Create Streamlined Permitting for Data Centers | 0.12 | 0.29 | 0.07 | 0.20 | 0.06 | 0.06 | 0.04 | 0.08 | 0.02 | 0.06 |
| AAP - Support Next Generation Manufacturing | 0.14 | 0.11 | 0.07 | 0.26 | 0.09 | 0.06 | 0.05 | 0.12 | 0.05 | 0.05 |
| AAP - Advance the Science of AI | 0.17 | 0.06 | 0.06 | 0.28 | 0.06 | 0.10 | 0.07 | 0.11 | 0.05 | 0.04 |
| AAP - Remove Red Tape and Onerous Regulation | 0.10 | 0.12 | 0.08 | 0.25 | 0.08 | 0.07 | 0.05 | 0.10 | 0.03 | 0.12 |
| AAP - Invest in AI Interpretability | 0.13 | 0.11 | 0.09 | 0.26 | 0.08 | 0.09 | 0.07 | 0.10 | 0.03 | 0.05 |
| AAP - Build an AI Evaluations Ecosystem | 0.10 | 0.09 | 0.06 | 0.30 | 0.07 | 0.10 | 0.06 | 0.13 | 0.03 | 0.06 |
| AAP - Develop a Grid to Match the Pace of AI Innovation | 0.09 | 0.12 | 0.13 | 0.22 | 0.06 | 0.06 | 0.04 | 0.14 | 0.01 | 0.12 |
| AAP - Invest in AI Enabled Science | 0.12 | 0.13 | 0.07 | 0.26 | 0.06 | 0.06 | 0.04 | 0.11 | 0.03 | 0.12 |
| AAP - Accelerate AI Adoption in Government | 0.15 | 0.13 | 0.10 | 0.23 | 0.03 | 0.10 | 0.02 | 0.11 | 0.02 | 0.11 |
| AAP - Encourage Open Source | 0.12 | 0.13 | 0.10 | 0.22 | 0.03 | 0.10 | 0.02 | 0.13 | 0.02 | 0.12 |
| AAP - Empower American Workers in the Age of AI | 0.14 | 0.07 | 0.08 | 0.24 | 0.05 | 0.06 | 0.04 | 0.28 | 0.02 | 0.03 |
| AAP - Enable AI Adoption | 0.12 | 0.13 | 0.10 | 0.23 | 0.03 | 0.10 | 0.02 | 0.13 | 0.02 | 0.11 |
| AAP - Restore American Semiconductor Manufacturing | 0.09 | 0.11 | 0.12 | 0.24 | 0.07 | 0.07 | 0.05 | 0.16 | 0.01 | 0.07 |
| AAP - Build Worldclass Data | 0.17 | 0.11 | 0.14 | 0.24 | 0.05 | 0.12 | 0.03 | 0.08 | 0.01 | 0.05 |
| AAP - Train a Skilled Workforce for AI Infrastructure | 0.10 | 0.08 | 0.13 | 0.21 | 0.07 | 0.09 | 0.07 | 0.18 | 0.02 | 0.05 |
| AAP - Export American AI to Allies and Partners | 0.14 | 0.10 | 0.10 | 0.23 | 0.05 | 0.06 | 0.03 | 0.20 | 0.02 | 0.04 |
| Totals | 3.90 | 3.29 | 3.06 | 5.63 | 1.91 | 2.59 | 1.33 | 3.59 | 0.93 | 1.76 |
| Average | 0.14 | 0.12 | 0.11 | 0.20 | 0.07 | 0.09 | 0.05 | 0.13 | 0.03 | 0.06 |
| Rank | 2 | 4 | 5 | 1 | 7 | 6 | 9 | 3 | 10 | 8 |

Note: AAP used for AI Action Plan.

*Table 9 Document Topic Distribution (National Security n=19)*

| Document | T1 | T2 | T3 | T4 | T5 | T6 | T7 | T8 | T9 | T10 |
|---|---|---|---|---|---|---|---|---|---|---|
| AI Diffusion Framework | 0.16 | 0.16 | 0.07 | 0.13 | 0.06 | 0.14 | 0.11 | 0.07 | 0.07 | 0.04 |
| AI Overwatch Act | 0.04 | 0.11 | 0.10 | 0.08 | 0.05 | 0.07 | 0.03 | 0.32 | 0.10 | 0.10 |
| Protecting AI and Cloud Competition in Defense Act | 0.05 | 0.06 | 0.40 | 0.13 | 0.02 | 0.13 | 0.02 | 0.08 | 0.02 | 0.10 |
| Advanced AI Security Readiness Act | 0.05 | 0.04 | 0.33 | 0.15 | 0.03 | 0.05 | 0.02 | 0.10 | 0.03 | 0.21 |
| Intelligence Authorization Act for FY26 | 0.05 | 0.04 | 0.34 | 0.15 | 0.03 | 0.05 | 0.02 | 0.10 | 0.03 | 0.20 |
| EAR - Integrated Circuits | 0.08 | 0.10 | 0.03 | 0.05 | 0.06 | 0.07 | 0.08 | 0.12 | 0.41 | |
| EAR – AI Model Weight Controls | 0.05 | 0.08 | 0.07 | 0.09 | 0.06 | 0.07 | 0.02 | 0.23 | 0.08 | 0.26 |
| NDAA 2026 | 0.05 | 0.06 | 0.37 | 0.13 | 0.03 | 0.07 | 0.02 | 0.11 | 0.02 | 0.13 |
| Semi Tariffs – Trump January 2026 | 0.05 | 0.04 | 0.35 | 0.13 | 0.03 | 0.06 | 0.02 | 0.10 | 0.03 | 0.20 |
| AI Model Weights FDPR | 0.04 | 0.09 | 0.07 | 0.09 | 0.06 | 0.08 | 0.02 | 0.21 | 0.07 | 0.27 |
| Protection Against Foreign Adversarial AI Act | 0.06 | 0.06 | 0.35 | 0.12 | 0.03 | 0.07 | 0.02 | 0.11 | 0.03 | 0.15 |
| GAIN AI Act | 0.05 | 0.11 | 0.10 | 0.08 | 0.05 | 0.07 | 0.03 | 0.31 | 0.10 | 0.10 |
| License Exception Artificial Intelligence Authorization (AIA) | 0.07 | 0.06 | 0.14 | 0.10 | 0.06 | 0.07 | 0.04 | 0.27 | 0.08 | 0.11 |
| Remote Access Security Act | 0.05 | 0.11 | 0.10 | 0.08 | 0.05 | 0.07 | 0.03 | 0.32 | 0.10 | 0.10 |
| AAP – Counter Chinese Influence in International Governance Bodies | 0.04 | 0.03 | 0.21 | 0.18 | 0.03 | 0.05 | 0.02 | 0.17 | 0.06 | 0.20 |
| AAP – Drive Adoption of AI within the Department of Defense | 0.05 | 0.05 | 0.36 | 0.14 | 0.03 | 0.06 | 0.02 | 0.09 | 0.02 | 0.19 |
| AAP – Plug Loopholes in Existing Semiconductor Manufacturing Export Controls | 0.06 | 0.05 | 0.33 | 0.13 | 0.03 | 0.07 | 0.02 | 0.11 | 0.03 | 0.17 |
| AAP – Align Protection Measures Globally | 0.06 | 0.06 | 0.20 | 0.27 | 0.04 | 0.07 | 0.02 | 0.09 | 0.04 | 0.15 |
| AAP – Strengthen AI Compute Export Control Enforcement | 0.06 | 0.05 | 0.31 | 0.13 | 0.03 | 0.07 | 0.02 | 0.13 | 0.03 | 0.17 |
| Totals | 1.12 | 1.63 | 3.58 | 2.73 | 1.14 | 1.83 | 0.51 | 2.82 | 1.16 | 2.46 |
| Average | 0.06 | 0.09 | 0.19 | 0.14 | 0.06 | 0.10 | 0.03 | 0.15 | 0.06 | 0.13 |
| Rank | 9 | 6 | 1 | 3 | 8 | 5 | 10 | 2 | 7 | 4 |

*Table 10 Document Topic Distribution (Civil & human rights n=14)*

| Document | T1 | T2 | T3 | T4 | T5 | T6 | T7 | T8 | T9 | T10 |
|---|---|---|---|---|---|---|---|---|---|---|
| EO 14319: Preventing Woke AI in the Federal Government | 0.08 | 0.07 | 0.13 | 0.08 | 0.05 | 0.03 | 0.05 | 0.39 | 0.07 | 0.04 |
| Utah Artificial Intelligence Policy Act | 0.07 | 0.11 | 0.10 | 0.04 | 0.19 | 0.07 | 0.07 | 0.06 | 0.04 | 0.25 |
| California Health and Insurance Code Amendment | 0.06 | 0.05 | 0.10 | 0.04 | 0.08 | 0.04 | 0.04 | 0.08 | 0.47 | 0.03 |
| Texas Responsible AI Governance Act | 0.10 | 0.11 | 0.30 | 0.06 | 0.11 | 0.07 | 0.07 | 0.09 | 0.06 | 0.04 |
| Illinois Human Rights Act | 0.14 | 0.11 | 0.11 | 0.25 | 0.10 | 0.06 | 0.06 | 0.09 | 0.07 | 0.03 |
| TAKE IT DOWN Act | 0.06 | 0.10 | 0.09 | 0.03 | 0.09 | 0.12 | 0.39 | 0.05 | 0.03 | 0.03 |
| California Generative AI: Training Data Transparency Bill | 0.11 | 0.07 | 0.10 | 0.07 | 0.37 | 0.06 | 0.07 | 0.06 | 0.06 | 0.04 |
| Maine Chatbot Disclosure Act | 0.08 | 0.19 | 0.11 | 0.03 | 0.10 | 0.04 | 0.08 | 0.05 | 0.02 | 0.27 |
| Colorado AI Act | 0.42 | 0.07 | 0.13 | 0.04 | 0.08 | 0.03 | 0.06 | 0.07 | 0.04 | 0.05 |
| NYC Automated Employment Decision Tools | 0.06 | 0.04 | 0.08 | 0.12 | 0.04 | 0.05 | 0.02 | 0.28 | 0.09 | 0.22 |
| TN ELVIS Act of 2024 | 0.11 | 0.07 | 0.39 | 0.07 | 0.05 | 0.04 | 0.03 | 0.06 | 0.02 | 0.15 |
| Florida Artificial Intelligence Bill of Rights | 0.07 | 0.04 | 0.10 | 0.27 | 0.05 | 0.05 | 0.02 | 0.08 | 0.03 | 0.29 |
| AAP - Ensure that Frontier AI Protects Free Speech and American Values | 0.14 | 0.05 | 0.15 | 0.05 | 0.04 | 0.02 | 0.03 | 0.41 | 0.06 | 0.03 |
| AAP - Combat Synthetic Media in the Legal System | 0.09 | 0.08 | 0.12 | 0.07 | 0.04 | 0.02 | 0.05 | 0.44 | 0.08 | 0.03 |
| Totals | 1.50 | 1.49 | 1.69 | 1.35 | 1.52 | 1.18 | 1.23 | 1.98 | 1.12 | 0.95 |
| Average | 0.11 | 0.11 | 0.12 | 0.10 | 0.11 | 0.08 | 0.09 | 0.14 | 0.08 | 0.07 |
| Rank | 4 | 5 | 2 | 6 | 3 | 8 | 7 | 1 | 9 | 10 |

*Table 11 Document Topic Distribution (Safety & Security n=20)*

| Document | T1 | T2 | T3 | T4 | T5 | T6 | T7 | T8 | T9 | T10 |
|---|---|---|---|---|---|---|---|---|---|---|
| EO 14365: Ensuring a National Policy Framework for Artificial Intelligence | 0.14 | 0.10 | 0.06 | 0.08 | 0.05 | 0.09 | 0.13 | 0.16 | 0.10 | 0.08 |
| EO 14144: Strengthening and Promoting Innovation in the Nation's Cybersecurity | 0.14 | 0.10 | 0.06 | 0.08 | 0.05 | 0.09 | 0.13 | 0.16 | 0.10 | 0.08 |
| California 2025 SB53 | 0.14 | 0.10 | 0.06 | 0.08 | 0.05 | 0.09 | 0.13 | 0.16 | 0.10 | 0.08 |
| California Generative AI: Training Data Transparency Bill | 0.13 | 0.09 | 0.06 | 0.08 | 0.05 | 0.09 | 0.14 | 0.16 | 0.09 | 0.08 |
| NY RAISE Act | 0.13 | 0.10 | 0.06 | 0.08 | 0.05 | 0.09 | 0.13 | 0.17 | 0.10 | 0.08 |
| Utah Artificial Intelligence Policy Act | 0.13 | 0.10 | 0.06 | 0.08 | 0.05 | 0.09 | 0.13 | 0.17 | 0.10 | 0.08 |
| EU AI Act | 0.14 | 0.10 | 0.06 | 0.08 | 0.05 | 0.09 | 0.13 | 0.16 | 0.10 | 0.08 |
| NIST AI RMF | 0.14 | 0.10 | 0.06 | 0.08 | 0.05 | 0.09 | 0.13 | 0.16 | 0.10 | 0.08 |
| ISO - AI Management System | 0.14 | 0.10 | 0.06 | 0.08 | 0.05 | 0.09 | 0.13 | 0.16 | 0.10 | 0.08 |
| AI Research, Innovation, and Accountability Act | 0.14 | 0.10 | 0.06 | 0.08 | 0.05 | 0.09 | 0.13 | 0.16 | 0.10 | 0.08 |
| Secure AI Act | 0.14 | 0.10 | 0.06 | 0.08 | 0.05 | 0.09 | 0.13 | 0.16 | 0.10 | 0.08 |
| AI Whistleblower Protection Act | 0.14 | 0.10 | 0.06 | 0.08 | 0.05 | 0.09 | 0.13 | 0.16 | 0.10 | 0.08 |
| AI Accountability Act | 0.14 | 0.10 | 0.06 | 0.08 | 0.05 | 0.09 | 0.13 | 0.16 | 0.10 | 0.08 |
| AAP - Bolster Critical Infrastructure Cybersecurity | 0.11 | 0.06 | 0.06 | 0.05 | 0.05 | 0.07 | 0.19 | 0.31 | 0.06 | 0.04 |
| AAP - Build High Security Data Centers | 0.09 | 0.05 | 0.08 | 0.06 | 0.09 | 0.12 | 0.15 | 0.23 | 0.08 | 0.06 |
| AAP - Ensure that the US is at the Forefront | 0.09 | 0.08 | 0.06 | 0.05 | 0.07 | 0.10 | 0.17 | 0.24 | 0.07 | 0.05 |
| AAP - Invest in Biosecurity | 0.09 | 0.05 | 0.05 | 0.06 | 0.11 | 0.08 | 0.17 | 0.26 | 0.11 | 0.04 |
| AAP - Promote Mature Federal Capacity for AI Incident Response | 0.10 | 0.07 | 0.05 | 0.05 | 0.06 | 0.08 | 0.22 | 0.29 | 0.04 | 0.05 |
| AAP - Promote Secure-By-Design AI | 0.09 | 0.05 | 0.07 | 0.06 | 0.09 | 0.12 | 0.15 | 0.23 | 0.08 | 0.06 |
| AAP - Protect Commercial and Government AI Innovations | 0.09 | 0.05 | 0.05 | 0.06 | 0.10 | 0.08 | 0.17 | 0.26 | 0.11 | 0.04 |
| Totals | 2.79 | 2.13 | 1.41 | 1.36 | 1.55 | 1.90 | 2.54 | 3.23 | 1.69 | 1.40 |
| Average | 0.14 | 0.11 | 0.07 | 0.07 | 0.08 | 0.10 | 0.13 | 0.16 | 0.08 | 0.07 |
| Rank | 2 | 4 | 8 | 10 | 7 | 5 | 3 | 1 | 6 | 9 |

Note for Tables 12 to 16: Each cell is the cosine similarity between the policy in the line and the rubric variable (line item) in the column.

*Table 12 Cosine Similarity – Policy vs Rubric Variables (Antitrust)*

| Document | Market Concentration & Dominance | Merger & Acquisition Scrutiny | Competition Enforcement Capacity | Interoperability & Open Standards | Data Access & Mandatory Sharing | Exclusionary Conduct & Platform Neutrality |
|---|---|---|---|---|---|---|
| Open App Markets Act | 47.93 | 44.94 | 55.08 | 56.17 | 50.61 | 48.56 |
| UK DMCCA 2024 | 53.96 | 52.85 | 53.12 | 45.32 | 48.77 | 51.02 |
| Platform Competition and Opportunity Act | 53.21 | 53.08 | 60.92 | 47.80 | 55.97 | 55.94 |
| Clayton Act | 48.14 | 48.01 | 51.55 | 43.45 | 42.15 | 51.29 |
| California Cartwright Act | 49.00 | 41.12 | 47.93 | 47.51 | 48.77 | 51.65 |
| Ending Platform Monopolies Act | 51.37 | 41.40 | 53.01 | 47.32 | 49.69 | 56.65 |
| EU DMA | 50.90 | 46.04 | 46.76 | 56.29 | 54.04 | 52.55 |
| Preventing Algorithmic Collusion Act | 55.76 | 52.81 | 61.03 | 52.74 | 56.57 | 62.11 |
| Sherman Act | 42.23 | 41.12 | 45.01 | 36.08 | 36.37 | 44.03 |
| CT Bill No. 8002 | 42.17 | 39.85 | 47.26 | 44.21 | 48.32 | 45.78 |

*Table 13 Cosine Similarity – Policy vs Rubric Variables (Industrial Policy)*

| Document | Workforce Development & Job Quality | R&D, Investment & Incentives | Domestic Manufacturing & Supply Chains | Infrastructure & Compute Capacity | Energy & Resource Capacity | Global Competitiveness |
|---|---|---|---|---|---|---|
| EO 14141: Advancing US Leadership in AI Infrastructure | 56.39 | 58.41 | 60.43 | 64.33 | 67.35 | 54.55 |
| EO 14275: Restoring Common Sense Federal Procurement | 35.00 | 39.97 | 38.57 | 35.69 | 35.69 | 36.92 |
| EO 14277: Advancing AI Edu. for American Youth | 59.00 | 54.90 | 47.65 | 49.86 | 45.31 | 53.27 |
| EO 14318: Accelerating Federal Permitting of Data Center Infrastructure | 46.02 | 50.59 | 57.60 | 61.35 | 63.50 | 45.30 |
| EO 14320: American AI Tech Stack | 46.80 | 58.59 | 53.98 | 49.20 | 45.38 | 64.59 |
| CREATE AI Act | 51.06 | 59.93 | 49.72 | 60.58 | 55.33 | 51.90 |
| EU AI Continent Action Plan | 54.05 | 54.86 | 46.32 | 56.27 | 55.32 | 52.08 |
| CHIPS and Science Act | 45.83 | 54.78 | 63.56 | 49.55 | 45.61 | 48.77 |
| California 2025 SB53 | 52.59 | 52.39 | 47.98 | 59.86 | 52.85 | 48.60 |
| AI and Critical Technology Workforce Framework Act | 56.75 | 53.84 | 50.76 | 52.34 | 46.98 | 57.09 |
| AI Workforce PREPARE Act | 63.61 | 56.46 | 50.16 | 53.69 | 48.77 | 53.73 |
| EU Apply AI Strategy | 55.96 | 49.88 | 48.74 | 49.43 | 51.44 | 51.98 |
| AAP - Create Streamlined Permitting for Data Centers | 41.96 | 49.56 | 52.71 | 56.49 | 57.86 | 45.37 |
| AAP - Support Next Generation Manufacturing | 53.25 | 63.01 | 64.92 | 52.00 | 48.28 | 55.51 |
| AAP - Advance the Science of AI | 48.37 | 59.02 | 48.23 | 50.61 | 44.83 | 52.64 |
| AAP - Remove Red Tape and Onerous Regulation | 43.20 | 52.02 | 41.82 | 43.13 | 41.12 | 51.00 |
| AAP - Invest in AI Interpretability | 43.70 | 53.74 | 47.68 | 44.82 | 37.48 | 50.51 |
| AAP - Build an AI Evaluations Ecosystem | 51.28 | 61.46 | 50.34 | 53.16 | 46.02 | 53.35 |
| AAP - Develop a Grid to Match the Pace of AI Innovation | 39.92 | 42.92 | 44.15 | 47.51 | 55.01 | 41.49 |
| AAP - Invest in AI Enabled Science | 45.18 | 57.21 | 45.89 | 53.31 | 45.47 | 43.33 |
| AAP - Accelerate AI Adoption in Government | 50.18 | 50.60 | 42.51 | 45.57 | 40.74 | 46.13 |
| AAP - Encourage Open Source | 42.40 | 54.49 | 42.38 | 52.71 | 44.24 | 46.19 |
| AAP - Empower American Workers in the Age of AI | 62.11 | 57.97 | 45.07 | 45.79 | 41.13 | 47.76 |
| AAP - Enable AI Adoption | 43.02 | 53.18 | 48.38 | 44.67 | 40.05 | 54.06 |
| AAP - Restore American Semiconductor Manufacturing | 41.68 | 48.39 | 60.51 | 42.55 | 38.51 | 47.26 |
| AAP - Build Worldclass Data | 43.37 | 49.34 | 44.52 | 49.29 | 43.36 | 49.15 |
| AAP - Train a Skilled Workforce for AI Infrastructure | 57.25 | 49.82 | 47.40 | 51.40 | 47.34 | 44.89 |
| AAP - Export American AI to Allies and Partners | 43.35 | 50.44 | 49.98 | 43.49 | 40.35 | 61.88 |

*Table 14 Cosine Similarity – Policy vs Rubric Variables (National Security)*

| Document | Defense & Military Applications | Critical Infrastructure Protection | Export Controls & Tech Sovereignty | Foreign Influence & Espionage | Cyber Defense & Resilience | Interagency & International Coordination |
|---|---|---|---|---|---|---|
| AI Diffusion Framework | 58.07 | 51.44 | 62.18 | 56.53 | 58.73 | 53.64 |
| AI Overwatch Act | 43.65 | 40.54 | 48.68 | 45.15 | 41.71 | 42.94 |
| Protecting AI and Cloud Competition in Defense Act | 41.49 | 42.60 | 53.69 | 52.95 | 44.57 | 50.37 |
| Advanced AI Security Readiness Act | 44.63 | 41.03 | 59.91 | 51.58 | 44.41 | 56.37 |
| Intelligence Authorization Act for FY26 | 49.01 | 47.07 | 63.83 | 55.93 | 49.03 | 54.07 |
| EAR - Integrated Circuits | 33.12 | 33.32 | 43.89 | 31.15 | 31.89 | 32.54 |
| EAR – AI Model Weight Controls | 43.65 | 40.54 | 48.68 | 45.15 | 41.71 | 42.94 |
| NDAA 2026 | 63.71 | 57.68 | 55.59 | 58.50 | 64.54 | 60.93 |
| Semi Tariffs – Trump January 2026 | 38.46 | 45.30 | 49.88 | 43.06 | 42.88 | 44.16 |
| AI Model Weights FDPR | 40.53 | 34.05 | 51.29 | 43.04 | 34.57 | 39.91 |
| Protecting Against Foreign Adversarial AI Act | 47.79 | 44.19 | 52.37 | 58.64 | 47.80 | 60.25 |
| GAIN AI Act | 45.01 | 42.60 | 53.69 | 52.95 | 44.57 | 50.37 |
| License Exception Artificial Intelligence Authorization (AIA) | 43.02 | 42.22 | 57.20 | 46.63 | 42.66 | 45.51 |
| Remote Access Security Act | 43.65 | 40.54 | 48.68 | 45.15 | 41.71 | 42.94 |
| AAP - Strengthen AI Compute Export Control Enforcement | 51.16 | 49.84 | 66.87 | 57.19 | 53.94 | 55.93 |
| AAP - Plug Loopholes in Existing Semiconductor Manufacturing Export Controls | 41.06 | 41.82 | 57.68 | 44.59 | 42.02 | 45.45 |
| AAP - Counter Chinese Influence in International Governance Bodies | 47.79 | 44.19 | 52.37 | 58.64 | 47.80 | 60.25 |
| AAP - Align Protection Measures Globally | 44.65 | 41.03 | 59.91 | 51.58 | 44.41 | 56.37 |
| AAP - Drive Adoption of AI within the Department of Defense | 56.63 | 43.26 | 39.38 | 42.00 | 48.92 | 46.02 |

*Table 15 Cosine Similarity – Policy vs Rubric Variables (Civil & human rights)*

| Document | Consumer Protection | Privacy & Data Rights | Bias & Fairness | Accessibility & Inclusion | Legal Protection & Due Process | Transparency & Explain-ability |
|---|---|---|---|---|---|---|
| EO 14319: Preventing Woke AI in the Federal Government | 46.31 | 45.82 | 53.18 | 50.54 | 47.79 | 50.87 |
| Utah Artificial Intelligence Policy Act | 69.27 | 57.36 | 57.11 | 58.88 | 59.60 | 60.10 |
| California Health and Insurance Code Amendment | 50.58 | 50.13 | 51.33 | 50.99 | 49.84 | 49.22 |
| Texas Responsible AI Governance Act | 63.99 | 58.90 | 59.10 | 56.71 | 55.80 | 56.41 |
| Illinois Human Rights Act | 57.05 | 52.24 | 58.39 | 57.25 | 53.63 | 55.24 |
| TAKE IT DOWN Act | 52.93 | 47.02 | 43.62 | 48.32 | 45.02 | 45.48 |
| California Generative AI: Training Data Transparency Bill | 52.68 | 59.27 | 48.79 | 51.81 | 49.83 | 55.35 |
| Maine Chatbot Disclosure Act | 58.18 | 42.03 | 43.15 | 45.46 | 45.05 | 50.66 |
| Colorado AI Act | 72.53 | 62.06 | 72.37 | 64.32 | 67.01 | 63.55 |
| NYC Automated Employment Decision Tools | 43.29 | 42.80 | 55.23 | 45.21 | 49.79 | 47.46 |
| TN ELVIS Act of 2024 | 47.76 | 45.67 | 43.21 | 45.08 | 45.92 | 45.64 |
| Florida Artificial Intelligence Bill of Rights | 58.54 | 59.48 | 53.31 | 54.69 | 58.80 | 54.57 |
| AAP - Ensure that Frontier AI Protects Free Speech and American Values | 52.14 | 48.34 | 51.36 | 53.37 | 52.65 | 49.41 |
| AAP - Combat Synthetic Media in the Legal System | 49.13 | 38.80 | 43.00 | 40.38 | 48.02 | 41.64 |

*Table 16 Cosine Similarity – Policy vs Rubric Variables (Safety & security)*

| Document | Testing & Evaluation Requirements | Incident Reporting & Transparency | Cybersec. Provisions | Human Oversight & Accountability | Whistlebl. Protections | Third-Party Audits & Compliance | Physical & Individual Safety Protections |
|---|---|---|---|---|---|---|---|
| EO 14365: Ensuring a National Policy Framework for Artificial Intelligence | 46.65 | 52.75 | 53.38 | 46.99 | 50.14 | 43.68 | 52.49 |
| EO 14144: Strengthening and Promoting Innovation in the Nation's Cybersecurity | 45.61 | 47.95 | 55.85 | 47.74 | 47.96 | 46.67 | 50.15 |
| California 2025 SB53 | 55.70 | 62.73 | 53.92 | 52.60 | 58.28 | 47.67 | 53.76 |
| California Generative AI: Training Data Transparency Bill | 49.20 | 54.13 | 59.06 | 48.60 | 46.88 | 48.06 | 50.89 |
| NY RAISE Act | 53.73 | 55.85 | 56.49 | 51.19 | 53.46 | 45.47 | 54.24 |
| Utah Artificial Intelligence Policy Act | 54.18 | 53.63 | 53.09 | 52.27 | 54.96 | 48.70 | 56.29 |
| EU AI Act | 58.67 | 55.49 | 61.26 | 59.31 | 53.40 | 55.59 | 61.08 |
| NIST AI RMF | 63.85 | 57.04 | 61.66 | 60.09 | 53.08 | 55.65 | 60.51 |
| ISO - AI Management System | 64.74 | 62.92 | 59.72 | 65.65 | 52.30 | 60.52 | 60.38 |
| AI Research, Innovation, and Accountability Act | 60.16 | 61.02 | 59.25 | 59.96 | 54.86 | 52.72 | 56.17 |
| Secure AI Act | 54.86 | 62.51 | 63.22 | 53.30 | 59.60 | 50.75 | 65.12 |
| AI Whistleblower Protection Act | 52.78 | 60.09 | 61.31 | 54.79 | 74.73 | 48.26 | 63.06 |
| AI Accountability Act | 50.56 | 54.93 | 51.78 | 59.51 | 49.54 | 51.46 | 49.97 |
| AAP - Bolster Critical Infrastructure Cybersecurity | 45.49 | 45.88 | 60.03 | 44.63 | 47.81 | 39.50 | 50.70 |
| AAP - Build High Security Data Centers | 44.45 | 43.08 | 54.61 | 40.23 | 42.92 | 38.18 | 45.28 |
| AAP - Ensure that the US is at the Forefront | 50.83 | 42.02 | 53.63 | 40.14 | 42.73 | 40.92 | 47.13 |
| AAP - Invest in Biosecurity | 45.90 | 46.07 | 54.00 | 43.93 | 48.98 | 40.57 | 50.36 |
| AAP - Promote Mature Federal Capacity for AI Incident Response | 46.02 | 53.05 | 51.44 | 46.34 | 46.26 | 38.78 | 46.24 |
| AAP - Promote Secure-By-Design AI | 56.13 | 53.02 | 64.05 | 54.62 | 54.52 | 49.86 | 56.17 |
| AAP - Protect Commercial and Government AI Innovations | 43.95 | 45.52 | 53.42 | 45.08 | 51.43 | 39.50 | 51.21 |

*Table 17 Dominant Policy Attribute per Document*

| Primary Attribute | Document | Antitrust | Industrial Policy | NatSec | Civil/Human Rights | Safety/Security |
|---|---|---|---|---|---|---|
| Antitrust | Open App Markets Act | **50.29** | 42.55 | 45.12 | 46.93 | 44.99 |
| Antitrust | UK DMCCA 2024 | **51.61** | 36.86 | 38.58 | 42.87 | 36.60 |
| Antitrust | Platform Competition and Opportunity Act | **54.33** | 39.94 | 40.08 | 45.64 | 38.66 |
| Antitrust | Clayton Act | **47.65** | 38.45 | 39.85 | 40.82 | 33.85 |
| Antitrust | California Cartwright Act | **47.10** | 37.13 | 36.21 | 42.85 | 35.25 |
| Antitrust | Ending Platform Monopolies Act | **49.19** | 37.17 | 38.35 | 43.00 | 37.72 |
| Antitrust | EU DMA | **50.31** | 39.24 | 42.79 | 46.00 | 44.43 |
| Antitrust | Preventing Algorithmic Collusion Act | **56.17** | 43.23 | 44.48 | 49.06 | 42.67 |
| Antitrust | Sherman Act | **41.15** | 36.14 | 40.71 | 38.80 | 36.51 |
| Antitrust | CT Bill No. 8002 | **43.83** | 39.19 | 35.71 | 40.02 | 35.31 |
| Industrial Policy | EO 14141: Advancing US Leadership in AI Infrastructure | 49.59 | **59.70** | 55.32 | 50.51 | 49.79 |
| Industrial Policy | EO 14275: Restoring Common Sense to Federal Procurement | 37.06 | 37.07 | **37.16** | 35.84 | 34.75 |
| Industrial Policy | EO 14277: Advancing AI Education for American Youth | 40.85 | **52.60** | 47.98 | 42.55 | 40.56 |
| Industrial Policy | EO 14318: Accelerating Federal Permitting of Data Center Infrastructure | 44.14 | **53.23** | 48.51 | 42.89 | 41.40 |
| Industrial Policy | EO 14320: American AI Tech Stack | 45.98 | **52.59** | 50.86 | 42.88 | 41.42 |
| Industrial Policy | CREATE AI Act | 48.60 | **54.51** | 48.28 | 46.58 | 45.38 |
| Industrial Policy | EU AI Continent Action Plan | 48.23 | **52.93** | 47.05 | 47.63 | 45.80 |
| Industrial Policy | CHIPS and Science Act | 41.63 | **52.00** | 47.83 | 42.04 | 42.67 |
| Industrial Policy | California 2025 SB53 | 44.86 | **52.35** | 46.73 | 47.89 | 43.87 |
| Industrial Policy | AI and Critical Technology Workforce Framework Act | 46.79 | **53.37** | 51.39 | 45.28 | 46.02 |
| Industrial Policy | AI Workforce PREPARE Act | 49.83 | **55.53** | 49.55 | 49.21 | 46.84 |
| Industrial Policy | EU Apply AI Strategy | 44.94 | 51.47 | **52.34** | 44.84 | 46.17 |
| Industrial Policy | AAP - Create Streamlined Permitting for Data Centers | 42.14 | **49.74** | 49.10 | 40.63 | 38.76 |
| Industrial Policy | AAP - Support Next Generation Manufacturing | 43.90 | **57.08** | 53.55 | 43.51 | 42.26 |
| Industrial Policy | AAP - Advance the Science of AI | 41.18 | **50.88** | 46.17 | 40.95 | 39.47 |
| Industrial Policy | AAP - Remove Red Tape and Onerous Regulation | **45.84** | 45.25 | 44.36 | 42.36 | 41.71 |
| Industrial Policy | AAP - Invest in AI Interpretability | 42.19 | 46.73 | **52.37** | 43.43 | 45.59 |
| Industrial Policy | AAP - Build an AI Evaluations Ecosystem | 45.04 | **53.09** | 47.15 | 43.28 | 48.46 |
| Industrial Policy | AAP - Develop a Grid to Match the Pace of AI Innovation | 35.87 | **44.17** | 41.07 | 31.89 | 30.89 |
| Industrial Policy | AAP - Invest in AI Enabled Science | 38.95 | **48.79** | 42.59 | 37.35 | 36.38 |
| Industrial Policy | AAP - Accelerate AI Adoption in Government | 42.32 | **46.69** | 45.08 | 43.13 | 41.39 |
| Industrial Policy | AAP - Encourage Open Source | 40.70 | **46.92** | 39.92 | 37.59 | 34.81 |
| Industrial Policy | AAP - Empower American Workers in the Age of AI | 39.62 | **51.89** | 41.77 | 40.14 | 37.36 |

*Continued on next page*

| Primary Attribute | Document | Antitrust | Industrial Policy | NatSec | Civil/Human Rights | Safety/Security |
|---|---|---|---|---|---|---|
| Industrial Policy | AAP - Enable AI Adoption | 43.07 | 47.18 | **51.08** | 40.65 | 41.64 |
| Industrial Policy | AAP - Restore American Semiconductor Manufacturing | 35.54 | **47.16** | 39.80 | 32.35 | 29.16 |
| Industrial Policy | AAP - Build Worldclass Data | 41.70 | **46.26** | 44.97 | 44.04 | 39.43 |
| Industrial Policy | AAP - Train a Skilled Workforce for AI Infrastructure | 35.31 | **50.69** | 40.51 | 33.84 | 32.78 |
| Industrial Policy | AAP - Export American AI to Allies and Partners | 44.23 | 47.67 | **50.50** | 40.24 | 38.52 |
| National Security | AI Diffusion Framework | 46.21 | 47.61 | **56.56** | 46.86 | 51.31 |
| National Security | AI Overwatch Act | 45.26 | 46.04 | **52.64** | 44.06 | 44.17 |
| National Security | Protecting AI and Cloud Competition in Defense Act | 53.12 | 54.70 | **56.36** | 50.43 | 47.54 |
| National Security | Advanced AI Security Readiness Act | 50.82 | 53.71 | **61.96** | 52.36 | 54.12 |
| National Security | Intelligence Authorization Act for FY26 | 49.73 | 51.81 | 53.74 | 50.48 | **54.46** |
| National Security | EAR - Integrated Circuits | 35.82 | 37.29 | **34.21** | 32.21 | 31.58 |
| National Security | EAR - AI Model Weight Controls | 43.14 | 44.02 | **43.61** | 42.55 | 43.91 |
| National Security | NDAA 2026 | 49.29 | 56.40 | **60.21** | 49.84 | 52.69 |
| National Security | Semi Tariffs - Trump January 2026 | 34.89 | 40.11 | **43.75** | 32.54 | 31.95 |
| National Security | AI Model Weights FDPR | 37.18 | 39.59 | **40.24** | 36.69 | 36.09 |
| National Security | Protecting Against Foreign Adversarial AI Act | 46.23 | 45.60 | **56.32** | 47.50 | 46.09 |
| National Security | GAIN AI Act | 45.43 | 47.98 | **50.71** | 45.06 | 42.89 |
| National Security | License Exception Artificial Intelligence Authorization (AIA) | 42.56 | 44.78 | **45.85** | 41.70 | 40.71 |
| National Security | Remote Access Security Act | 39.38 | 38.99 | **47.05** | 39.81 | 38.93 |
| National Security | AAP - Counter Chinese Influence in International Governance Bodies | 44.26 | 45.48 | **51.26** | 43.62 | 40.20 |
| National Security | AAP - Drive Adoption of AI within the Department of Defense | 36.38 | 44.58 | **46.43** | 36.41 | 36.14 |
| National Security | AAP - Plug Loopholes in Existing Semiconductor Manufacturing Export Controls | 37.67 | 42.11 | **45.04** | 33.92 | 33.59 |
| National Security | AAP - Align Protection Measures Globally | 37.82 | 39.43 | **48.98** | 36.39 | 35.26 |
| National Security | AAP - Strengthen AI Compute Export Control Enforcement | 47.99 | 51.51 | **55.29** | 44.43 | 43.99 |
| Civil & Human Rights | EO 14319: Preventing Woke AI in the Federal Government | 44.88 | 45.45 | 48.15 | **48.90** | 46.74 |
| Civil & Human Rights | Utah AI Policy Act SB0149 | 51.36 | 54.15 | 55.36 | **60.53** | 56.21 |
| Civil & Human Rights | California Health and Insurance Code Amendment | 44.72 | 45.01 | 46.59 | **50.45** | 47.15 |
| Civil & Human Rights | Texas Responsible AI Governance Act | 47.97 | 50.33 | 52.62 | **58.92** | 53.61 |
| Civil & Human Rights | Illinois Human Rights Act | 48.86 | 53.00 | 51.30 | **55.69** | 50.49 |
| Civil & Human Rights | TAKE IT DOWN Act | 43.95 | 40.16 | 42.93 | **47.26** | 40.81 |

*Continued on next page*

| Primary Attribute | Document | Antitrust | Industrial Policy | NatSec | Civil/Human Rights | Safety/Security |
|---|---|---|---|---|---|---|
| Civil & Human Rights | California Generative AI: Training Data Transparency Bill | 46.23 | 48.31 | 51.69 | **52.93** | 50.54 |
| Civil & Human Rights | Maine Chatbot Disclosure Act | 42.89 | 40.44 | 41.30 | **47.31** | 42.60 |
| Civil & Human Rights | Colorado AI Act | 53.83 | 51.58 | 56.69 | **67.45** | 60.69 |
| Civil & Human Rights | NYC Automated Employment Decision Tools | 38.94 | 38.06 | 36.49 | **47.26** | 42.31 |
| Civil & Human Rights | TN ELVIS Act of 2024 | 38.39 | 38.36 | 40.81 | **45.54** | 40.18 |
| Civil & Human Rights | Florida AI Bill of Rights | 49.39 | 49.36 | 51.52 | **56.75** | 49.89 |
| Civil & Human Rights | AAP - Ensure that Frontier AI Protects Free Speech and American Values | 47.94 | 48.86 | **51.87** | 51.21 | 48.72 |
| Civil & Human Rights | AAP - Combat Synthetic Media in the Legal System | 38.03 | 36.18 | **43.65** | 43.61 | 41.13 |
| Safety/Security | EO 14365: Ensuring a National Policy Framework for AI | 50.58 | 52.27 | 53.05 | **53.53** | 49.26 |
| Safety/Security | EO 14144: Strengthening and Promoting Innovation in the Nation's Cybersecurity | 46.82 | 47.60 | **55.76** | 49.96 | 48.61 |
| Safety/Security | California 2025 SB53 | 48.37 | 49.34 | 50.85 | 52.08 | **55.28** |
| Safety/Security | California Generative AI: Training Data Transparency Bill | 47.03 | 48.61 | 52.70 | **54.05** | 51.15 |
| Safety/Security | NY RAISE Act | 47.41 | 50.85 | 50.28 | 53.12 | **53.28** |
| Safety/Security | Utah AI Policy Act | 49.31 | 53.63 | 53.89 | **55.54** | 53.39 |
| Safety/Security | EU AI Act | 47.41 | 44.41 | 51.58 | 55.61 | **58.08** |
| Safety/Security | NIST AI RMF | 45.90 | 47.12 | 53.20 | 58.24 | **59.70** |
| Safety/Security | ISO - AI Management System | 47.90 | 49.98 | 52.72 | 56.31 | **61.75** |
| Safety/Security | AI Research, Innovation, and Accountability Act | 52.47 | 55.36 | 57.97 | 58.30 | **58.45** |
| Safety/Security | Secure AI Act | 47.99 | 50.94 | 57.70 | 53.42 | **58.12** |
| Safety/Security | AI Whistleblower Protection Act | 48.85 | 50.63 | 57.15 | 56.58 | **58.23** |
| Safety/Security | AI Accountability Act | 48.56 | 48.79 | 52.06 | 52.11 | **52.67** |
| Safety/Security | AAP - Bolster Critical Infrastructure Cybersecurity | 42.95 | 48.20 | **58.57** | 44.84 | 47.76 |
| Safety/Security | AAP - Build High Security Data Centers | 42.74 | 49.27 | **54.66** | 44.27 | 44.44 |
| Safety/Security | AAP - Ensure that the US is at the Forefront | 40.64 | 43.22 | **53.60** | 40.43 | 46.15 |
| Safety/Security | AAP - Invest in Biosecurity | 43.92 | 47.89 | **54.42** | 46.10 | 47.07 |
| Safety/Security | AAP - Promote Mature Federal Capacity for AI Incident Response | 42.54 | 45.51 | **52.26** | 43.06 | 47.26 |
| Safety/Security | AAP - Promote Secure-By-Design AI | 47.59 | 53.59 | **61.50** | 53.48 | 55.84 |
| Safety/Security | AAP - Protect Commercial and Government AI Innovations | 45.78 | 51.50 | **57.99** | 47.75 | 46.80 |

*Table 18 LLM Benchmark – Antitrust - Policy from the Model Policy Data Set*

**LLM Benchmark (Antitrust) - EU DMA (2025)**

| Model | Antitrust | Industrial Policy | NatSec | Civil & Human Rights | Safety/Security |
|---|---|---|---|---|---|
| Rubric-Calibrated Model | 50.31 | 39.24 | 42.79 | 46.00 | 44.43 |
| OpenAI GPT-5.2 | 78.75 | 1.0 | 4.0 | 29.0 | 28.25 |
| Claude Sonnet 4.5 | 91.5 | 4.17 | 3.75 | 33.33 | 12.5 |
| Google Gemini 3 Pro | 100 | 2.5 | 11.25 | 53.75 | 25 |
| Grok 4.1 Thinking | 100 | 0.0 | 3.75 | 60.25 | 13.75 |
| DeepSeek-V3.2 Thinking | 77.5 | 0.0 | 0.0 | 55.0 | 0.0 |
| Kimi K2.5 | 83.75 | 2.5 | 0.0 | 45.0 | 11.25 |

*Table 19 LLM Benchmark - Industrial Policy - Policy from the Model Policy Dataset*

**LLM Benchmark (Industrial Policy) - Executive Order 14320: Promoting the Export of the American AI Technology Stack**

| Model | Antitrust | Industrial Policy | NatSec | Civil & Human Rights | Safety/Security |
|---|---|---|---|---|---|
| Rubric-Calibrated Model | 45.98 | 52.59 | 50.86 | 42.88 | 41.42 |
| OpenAI GPT-5.2 | 3.75 | 35.0 | 33.25 | 0.0 | 3.75 |
| Claude Sonnet 4.5 | 0.0 | 50.0 | 26.25 | 0.0 | 3.75 |
| Google Gemini 3 Pro | 0.0 | 45.0 | 27.5 | 0.0 | 7.5 |
| Grok 4.1 Thinking | 0.0 | 30.0 | 27.5 | 0.0 | 3.75 |
| DeepSeek-V3.2 Thinking | 0.0 | 42.5 | 27.5 | 0.0 | 3.75 |
| Kimi K2.5 | 3.75 | 48.75 | 31.25 | 0.0 | 13.75 |

*Table 20 LLM Benchmark – National Security - Policy from the Model Policy Dataset*

**LLM Benchmark (National Security) - 2026 NDAA: Subtitle D - Artificial Intelligence**

| Model | Antitrust | Industrial Policy | NatSec | Civil & Human Rights | Safety/Security |
|---|---|---|---|---|---|
| Rubric-Calibrated Model | 49.29 | 56.40 | 60.21 | 49.84 | 52.69 |
| OpenAI GPT-5.2 | 7.25 | 28.25 | 55.75 | 7.0 | 46.0 |
| Claude Sonnet 4.5 | 7.5 | 37.5 | 62.5 | 20.83 | 43.75 |
| Google Gemini 3 Pro | 10.0 | 47.5 | 71.25 | 18.75 | 55.0 |
| Grok 4.1 Thinking | 0.0 | 24.0 | 59.0 | 5.0 | 19.0 |
| DeepSeek-V3.2 Thinking | 0.0 | 10.0 | 38.75 | 0 | 33.75 |
| Kimi K2.5 | 10.0 | 33.75 | 63.75 | 18.75 | 43.75 |

*Table 21 LLM Benchmark – Civil & human rights – Policy from the Model Policy Dataset*

**LLM Benchmark (Civil & Human Rights) - Colorado AI Act**

| Model | Antitrust | Industrial Policy | NatSec | Civil & Human Rights | Safety/Security |
|---|---|---|---|---|---|
| Rubric-Calibrated Model | 53.83 | 51.58 | 56.69 | 67.45 | 60.69 |
| OpenAI GPT-5.2 | 0.0 | 0.0 | 1.0 | 65.75 | 45.0 |
| Claude Sonnet 4.5 | 0.0 | 0.0 | 4.17 | 70.83 | 53.75 |
| Google Gemini 3 Pro | 0.0 | 0.0 | 3.75 | 87.5 | 50.0 |
| Grok 4.1 Thinking | 0.0 | 0.0 | 0.0 | 63.75 | 32.5 |
| DeepSeek-V3.2 Thinking | 0.0 | 0.0 | 0.0 | 81.25 | 37.5 |
| Kimi K2.5 | 0.0 | 3.0 | 9.0 | 71.0 | 36.0 |

*Table 22 LLM Benchmark – Safety & Security - Policy from the Model Policy Dataset*

**LLM Benchmark (Safety/Security) - California 2025 SB53**

| Model | Antitrust | Industrial Policy | NatSec | Civil & Human Rights | Safety/Security |
|---|---|---|---|---|---|
| Rubric-Calibrated Model | 48.37 | 49.34 | 50.85 | 52.08 | 55.28 |
| OpenAI GPT-5.2 | 0.0 | 15.5 | 8.25 | 18.0 | 62.75 |
| Claude Sonnet 4.5 | 0.0 | 12.5 | 12.5 | 12.5 | 60.75 |
| Google Gemini 3 Pro | 0.0 | 22.5 | 11.25 | 12.5 | 78.75 |
| Grok 4.1 Thinking | 0.0 | 6.25 | 0.0 | 10.0 | 67.5 |
| DeepSeek-V3.2 Thinking | 0.0 | 23.75 | 0.0 | 10.0 | 50.0 |
| Kimi K2.5 | 16.25 | 23.75 | 3.75 | 21.25 | 57.5 |

## References

Allen, D., Hubbard, S., Lim, W., Stanger, A., Wagman, S., Zalesne, K., & Omoakhalen, O. (2025). A roadmap for governing AI: technology governance and power-sharing liberalism. *Ai and Ethics (Online)*, *5*(3), 3355–3377. https://doi.org/10.1007/s43681-024-00635-y

Ashford, N. A., & Heaton, G. R. (1976). Environmental and safety regulations: Reasons for their adoption and possible effects on technological innovation. *Environmental Policy and Law*, *1*(4), 172–176. https://doi.org/10.1016/S0378-777X(76)80123-0

Bardach, E., & Patashnik, E. M. (2023). *A practical guide for policy analysis: The eightfold path to more effective problem solving*. CQ press.

Bengio, Y. (2026, February). *International AI Safety Report 2026 | International AI Safety Report*. https://internationalaisafetyreport.org/publication/international-ai-safety-report-2026

Bengio, Y., Mindermann, S., Privitera, D., Besiroglu, T., Bommasani, R., Casper, S., Choi, Y., Fox, P., Garfinkel, B., Goldfarb, D., Heidari, H., Ho, A., Kapoor, S., Khalatbari, L., Longpre, S., Manning, S., Mavroudis, V., Mazeika, M., Michael, J., … Zeng, Y. (2025). *International AI Safety Report*. arXiv.

Bengio, Y., Tegmark, M., Russell, S., Song, D., Mindermann, S., Xue, L., Casper, S., Ong, L., Wilfred, V., Maharaj, T., Lee, W. S., & Zhang, Y.-Q. (2025). The Singapore Consensus on Global AI Safety Research Priorities: Building a Trustworthy, Reliable and Secure AI Ecosystem. *SuperIntelligence - Robotics - Safety & Alignment*, *2*(5). https://doi.org/10.70777/si.v2i5.15503

Bick, A., Blandin, A., & Deming, D. (2024). *The Rapid Adoption of Generative AI* (No. W32966; p. w32966). National Bureau of Economic Research. https://doi.org/10.3386/w32966

Birhane, A., Kasirzadeh, A., Leslie, D., & Wachter, S. (2023). Science in the age of large language models. *Nature Reviews Physics*, *5*(5), 277–280. https://doi.org/10.1038/s42254-023-00581-4

Bradford, A. (2023). *Digital empires: The global battle to regulate technology* (1st ed.). Oxford University Press. https://doi.org/10.1093/oso/9780197649268.001.0001

Breyer, S. G. (1984). *Regulation and Its Reform*. Harvard University Press.

Brynjolfsson, E., Li, D., & Raymond, L. (2025). Generative AI at Work. *The Quarterly Journal of Economics*, *140*(2), 889–942. https://doi.org/10.1093/qje/qjae044

Brynjolfsson, E., & McAfee, A. (2016). *The second machine age: Work, progress, and prosperity in a time of brilliant technologies* (First published as a Norton paperback). W. W. Norton & Company.

Buyl, M., Khalaf, H., Mayrink Verdun, C., Monteiro Paes, L., Vieira Machado, C. C., & du Pin Calmon, F. (2025). Ai alignment at your discretion. *Proceedings of the 2025 ACM Conference on Fairness, Accountability, and Transparency*, 3046–3074.

Calvino, F., Haerle, D., & Liu, S. (2025). Is generative AI a General Purpose Technology?: Implications for productivity and policy. *OECD Artificial Intelligence Papers*. https://doi.org/10.1787/704e2d12-en

Campello, R. J., Moulavi, D., & Sander, J. (2013). Density-based clustering based on hierarchical density estimates. *Pacific-Asia Conference on Knowledge Discovery and Data Mining*, 160–172.

Carvão, P. (2024). The dual imperative: Innovation and regulation in the AI era. *International Journal of Technology Policy and Law*, *3*(3), 236–251. https://doi.org/10.1504/IJTPL.2024.142861

Carvão, P., Ancheva, S., Atir, Y., Jeloka, S., & Zhou, B. (2025). *Governance at a Crossroads: Artificial Intelligence and the Future of Innovation in America*. https://dash.harvard.edu/handle/1/42716416

Carvão, P., Atir, Y., & Ancheva, S. (2025). A Dynamic Governance Model for AI. *Lawfare*. https://www.lawfaremedia.org/article/a-dynamic-governance-model-for-ai

Carvão, P., Jeloka, S., & Yashiro, M. (2025). *The People Have Spoken: The Tech Industry, Civil Society, and the U.S. Artificial Intelligence Action Plan*. https://dash.harvard.edu/workflowitems/84107/view

Castañeira, J. A. E., Brando, A., Laukyte, M., & Serra-Vidal, M. (2025, October). Position: If Innovation in AI systematically Violates Fundamental Rights, Is It Innovation at All? *The Thirty-Ninth Annual Conference on Neural Information Processing Systems Position Paper Track*.

Center for AI Safety. (2023). *Statement on AI Risk | CAIS*. Center for AI Safety. Center for AI Safety. https://aistatement.com

Chander, A. (2014). How law made Silicon Valley. *Emory Law Journal*, *63*(3), 639–694.

Collingridge, D. (1980). *The social control of technology*. St. Martin's Press.

De Gregorio, G., & Dunn, P. (2022). The European risk-based approaches: Connecting constitutional dots in the digital age. *Common Market Law Review*, *59*(Issue 2), 473–500. https://doi.org/10.54648/COLA2022032

Derthick, M., Quirk, P. J., & Brookings Institution. (1985). *The politics of deregulation* (1st ed.). Brookings Institution.

Eloundou, T., Manning, S., Mishkin, P., & Rock, D. (2024). GPTs are GPTs: Labor market impact potential of LLMs. *Science (American Association for the Advancement of Science)*, *384*(6702), 1306–1308. https://doi.org/10.1126/science.adj0998

Engstrom, D. F., Ho, D. E., Sharkey, C. M., & Cuéllar, M.-F. (2020). Government by algorithm: Artificial intelligence in federal administrative agencies. *NYU School of Law, Public Law Research Paper*, (20–54).

Filimonovic, D., Rutzer, C., & Wunsch, C. (2025). *Can GenAI Improve Academic Performance? Evidence from the Social and Behavioral Sciences* (arXiv:2510.02408). arXiv. https://doi.org/10.48550/arXiv.2510.02408

Fischer, F., & Forester, J. (1993). *The argumentative turn in policy analysis and planning* (1st ed.). Duke University Press.

Glaeser, E., & Shleifer, A. (2003). *The Rise of the Regulatory State*. https://doi.org/10.1257/002205103765762725

Grimmer, J., & Stewart, B. M. (2013). Text as data: The promise and pitfalls of automatic content analysis methods for political texts. *Political Analysis*, *21*(3), 267–297.

Grootendorst, M. (2022). BERTopic: Neural topic modeling with a class-based TF-IDF procedure. *arXiv Preprint arXiv:2203.05794*.

G'sell, F. (2024). *Regulating Under Uncertainty: Governance Options for Generative AI*. Stanford Cyber Policy Center. https://cyber.fsi.stanford.edu/content/regulating-under-uncertainty-governance-options-generative-ai

Ha, H. (2025). When AI meets AI: analyzing AI bills using AI: The chamber and partisan effects in US AI Governance. *AI & SOCIETY*, 1–26.

Haidt, J. (2024). *The anxious generation: How the great rewiring of childhood is causing an epidemic of mental illness*. Penguin Press.

Healy, J., & McInnes, L. (2024). Uniform manifold approximation and projection. *Nature Reviews Methods Primers*, *4*(1), 82.

Hendrycks, D., Mazeika, M., & Woodside, T. (2023). *An Overview of Catastrophic AI Risks* (arXiv:2306.12001). arXiv. http://arxiv.org/abs/2306.12001

Hilliard, A., Gulley, A., Kazim, E., & Koshiyama, A. S. (2026). Artificial intelligence policy worldwide: A comparative analysis. *Royal Society Open Science*, *13*(2), 242234. https://doi.org/10.1098/rsos.242234

Honnibal, M., Montani, I., Van Landeghem, S., & Boyd, A. (2020). *spaCy: Industrial-strength natural language processing in python*.

Jin, Z., & Mihalcea, R. (2022). Natural language processing for policymaking. In *Handbook of computational social science for policy* (pp. 141–162). Springer International Publishing Cham.

Jobin, A., Ienca, M., & Vayena, E. (2019). The global landscape of AI ethics guidelines. *Nature Machine Intelligence*, *1*(9), 389–399. https://doi.org/10.1038/s42256-019-0088-2

Lee, F., & Ribes, D. (2025). Computational universalism, or, Attending to relationalities at scale. *Social Studies of Science*, 03063127251345089.

Lessig, L. (1999). *Code and other laws of cyberspace*. Basic Books.

Marchant, G. E., Allenby, B. R., & Herkert, J. R. (2011). *The growing gap between emerging technologies and legal-ethical oversight: The pacing problem* (1st ed., Vol. 7). Springer. https://doi.org/10.1007/978-94-007-1356-7

McLean, S., Read, G. J. M., Thompson, J., Baber, C., Stanton, N. A., & Salmon, P. M. (2023). The risks associated with Artificial General Intelligence: A systematic review. *Journal of Experimental & Theoretical Artificial Intelligence*, *35*(5), 649–663. https://doi.org/10.1080/0952813X.2021.1964003

Moses, L. B. (2007). Recurring dilemmas: The law's race to keep up with technological change. *U. Ill. JL Tech. & Pol'y*, 239.

Persily, N., & Tucker, J. A. (Eds.). (2020). Social Media and Democracy. In *Social Media and Democracy* (pp. i–i). Cambridge University Press. Cambridge Core. https://www.cambridge.org/core/product/29E8721497D280A0CEE5ACE838A7919A

Robert, C., & Zeckhauser, R. (2011). The methodology of normative policy analysis. *Journal of Policy Analysis and Management*, *30*(3), 613–643.

Sanders, N. E., & Schneier, B. (2025). *Rewiring democracy: How AI will transform our politics, government, and citizenship*. The MIT Press.

Sharma, S. (2024). Benefits or concerns of AI: A multistakeholder responsibility. *Futures*, *157*, 103328. https://doi.org/10.1016/j.futures.2024.103328

Sparck Jones, K. (1972). A statistical interpretation of term specificity and its application in retrieval. *Journal of Documentation*, *28*(1), 11–21.

Stigler, G. J. (1971). The Theory of Economic Regulation. *Bell Journal of Economics and Management Science*, *2*(1), 3–21.

Sunstein, C. R. (1990). Paradoxes of the Regulatory State. *The University of Chicago Law Review*, *57*(2), 407–441. https://doi.org/10.2307/1599951

Trajtenberg, M. (2019). *Artificial Intelligence as the Next GPT: A Political-Economy Perspective* (A. Agrawal, J. Gans, & A. Goldfarb, Eds.; pp. 175–186). University of Chicago Press. https://doi.org/10.7208/9780226613475-008

Tuan Pham, Evani Radiya-Dixit, Marissa Gerchick, Brooke Madubuonwu, & Suresh Venkatasubramanian. (2026, January 8). Using AI to Make Sense of AI Policy. *American Civil Liberties Union*. https://www.aclu.org/publications/using-ai-to-make-sense-of-ai-policy

Wang, S., Zhang, Y., Xiao, Y., & Liang, Z. (2025). Artificial intelligence policy frameworks in China, the European Union and the United States: An analysis based on structure topic model. *Technological Forecasting and Social Change*, *212*, 123971. https://doi.org/10.1016/j.techfore.2025.123971

Weimer, D. L. (2005). Institutionalizing neutrally competent policy analysis: Resources for promoting objectivity and balance in consolidating democracies. *Policy Studies Journal*, *33*(2), 131–146.

Weimer, D. L., & Vining, A. R. (2017). *Policy analysis: Concepts and practice*. Routledge.

Zhao, X., Hu, X., Shan, Z., Huang, S., Zhou, Y., Zhang, X., Sun, Z., Liu, Z., Li, D., & Wei, X. (2025). Kalm-embedding-v2: Superior training techniques and data inspire A versatile embedding model. *arXiv Preprint arXiv:2506.20923*.

Zhong, H., Xiao, C., Tu, C., Zhang, T., Liu, Z., & Sun, M. (2020). How does NLP benefit legal system: A summary of legal artificial intelligence. *arXiv Preprint arXiv:2004.12158*.